\documentclass[12pt,a4paper]{article}
\usepackage{multirow}
\usepackage{amsmath}
\usepackage{graphicx}
\usepackage[colorlinks=true,
    linkcolor=black,
    citecolor=black,
    urlcolor=black]{hyperref}
\usepackage{cite}
\usepackage{geometry}
\usepackage{tikz}
\usetikzlibrary{calc}
\usepackage{chngcntr}
\usepackage{float}  
\counterwithin{figure}{section}
\usepackage{tikz}
\usetikzlibrary{shapes.geometric, arrows.meta, positioning, calc, shadows, fit}

\begin{document}
\begin{titlepage}
\thispagestyle{empty} 
\begin{tikzpicture}[remember picture,overlay]
    \node at (current page.center) {
        \includegraphics[width=\paperwidth,height=\paperheight]{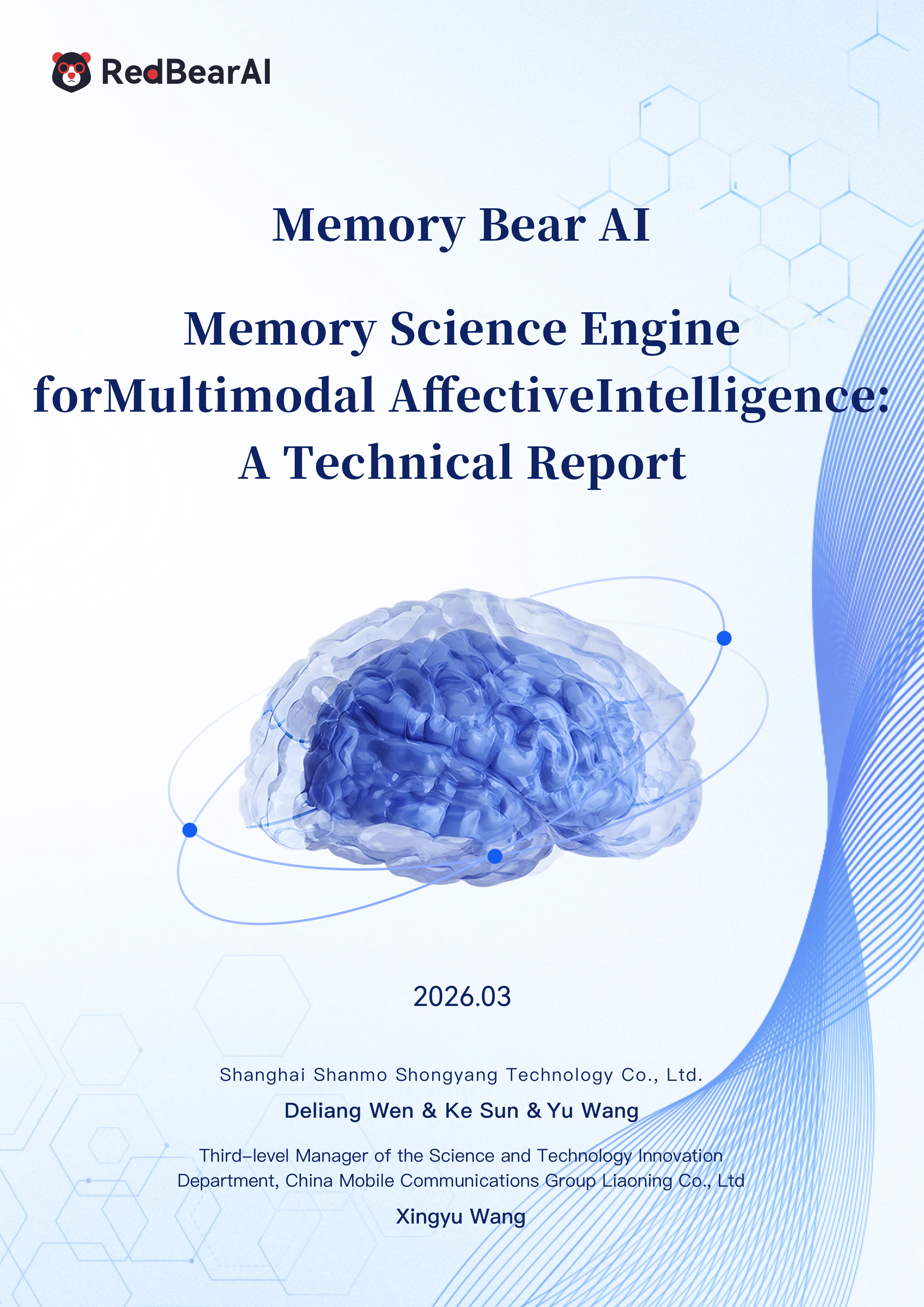}
    };
\end{tikzpicture}
\end{titlepage}


\begin{abstract}
Affective judgment in real interaction is rarely a purely local prediction problem. Emotional meaning often depends on prior trajectory, contextual accumulation, and multimodal evidence that may be weak, noisy, or incomplete at the current moment. Although recent progress in multimodal emotion recognition (MER) has improved the integration of textual, acoustic, and visual signals, many existing systems still remain optimized for short-range inference and provide limited support for persistent affective memory, long-horizon dependency modeling, and context-aware robustness under imperfect input.

This technical report presents the \textbf{Memory Bear AI Memory Science Engine}, a memory-centered framework for multimodal affective intelligence. Instead of treating emotion as a transient output label, the proposed framework models affective information as a structured and evolving variable within a memory system. The architecture organizes multimodal affective processing through structured memory formation, working-memory aggregation, long-term consolidation, memory-driven retrieval, dynamic fusion calibration, and continuous memory updating. At its core, the framework transforms multimodal signals into structured \emph{Emotion Memory Units} (EMUs), allowing affective information to be preserved, reactivated, and revised across interaction horizons.

Experimental results show that the proposed framework consistently outperforms comparison systems across benchmark and business-grounded settings. On IEMOCAP and CMU-MOSEI, Memory Bear AI achieves the strongest overall performance with accuracies of 78.8 and 66.7, respectively. On the Memory Bear AI Business Dataset, the framework reaches 68.4 accuracy, 48.6 weighted F1, and 45.9 macro F1, improving accuracy by 8.2 points over a traditional fusion baseline. Under degraded multimodal conditions, the framework also maintains the strongest robustness, preserving 92.3\% of complete-condition performance.

The report argues that the value of this design lies not only in stronger multimodal fusion, but in the ability to reuse historically relevant affective information when current evidence alone is insufficient. This enables more stable affective judgment under long-horizon interaction, noisy modalities, and missing-modality conditions. Through architectural analysis, experimental validation, and case-based discussion, the report positions the Memory Bear AI engine as a practical step from local emotion recognition toward more continuous, robust, and deployment-relevant affective intelligence.
\end{abstract}


\newpage
\tableofcontents

\newpage
\section{Introduction}

\subsection{Affective Judgment as a Memory-Centered Problem}

In real interaction, affective judgment is rarely a matter of reading off a single emotional label from the current moment. Emotional meaning is often weakly expressed, context-dependent, and distributed across time\cite{tyng2017emotionmemory,faul2022moodcongruent,kensinger2009details,clore2007judgment}. A short utterance may sound neutral in isolation but carry resignation when placed after repeated frustration; a restrained facial expression may only become interpretable when combined with prior conversational history; a noisy acoustic cue may be misleading unless evaluated against a longer emotional trajectory. In practice, affective judgment often depends not only on what is visible or audible now, but also on what has been previously experienced, retained, and reactivated\cite{tyng2017emotionmemory,faul2022moodcongruent,kensinger2009details,clore2007judgment}.

This observation motivates the central premise of this report: affective intelligence should be treated as a memory-centered problem rather than as a purely local classification problem. The challenge is not simply to recognize emotion from multimodal input, but to preserve, organize, retrieve, and update affective information across interaction horizons. From this perspective, memory is not an auxiliary component added after perception. It is part of the mechanism through which present emotional meaning becomes interpretable.

The present report develops this idea through the \textbf{Memory Bear AI Memory Science Engine}, a system architecture that brings structured memory mechanisms into multimodal affective judgment. Instead of treating emotional inference as a one-step mapping from current signals to output labels, the proposed framework models affective interpretation as a process in which multimodal evidence is encoded into structured memory, selectively retained, reactivated when contextually relevant, and revised as new evidence accumulates. The goal is therefore not only stronger local emotion prediction, but more persistent, robust, and context-sensitive affective understanding.

\subsection{Why Existing Multimodal Approaches Are Still Insufficient}

Multimodal emotion recognition has significantly improved affective modeling by combining textual, acoustic, and visual signals \cite{lian2024merbench,ramaswamy2024merreview}. This progress is important, because emotion is rarely expressed through a single channel. However, many existing multimodal systems still remain optimized for short-horizon prediction. Their primary objective is typically to infer the emotional state expressed in the current utterance, segment, or interaction window, even when stronger multimodal fusion is used.

This limitation becomes visible in three ways. First, many systems lack explicit temporal depth: they can model local context, but they do not systematically organize historical affective information into persistent memory. Second, emotion is often treated as an output label rather than as a variable that should continue to shape later interpretation. Third, robustness under missing modalities, unstable signal quality, and contextually ambiguous input remains limited, because present inference still depends heavily on the information available in the current local window \cite{li2024multimodalalignmentfusion,ramaswamy2024merreview,arunagladys2023survey}.

These limitations suggest that stronger multimodal fusion alone is not sufficient. What remains missing is a structured mechanism through which affective information can persist beyond the current moment and later influence interpretation when local evidence is weak, noisy, or incomplete.

\subsection{Memory Bear AI as a Memory-Centered Solution}

The Memory Bear AI framework addresses this gap by treating memory as the organizing
layer of multimodal affective judgment. Under this view, affective information is not
only perceived, but also encoded into structured memory units, aggregated in short-term
working memory, selectively consolidated into long-term memory, retrieved when
contextually relevant, and revised through continuous updating. This memory-centered
design is also consistent with the broader Memory Bear architecture, which frames
memory as a foundational mechanism for long-horizon continuity, knowledge fidelity,
and adaptive cognition rather than as a passive storage layer \cite{wen2025memorybear}.

The proposed approach is especially useful in three situations. The first is long-horizon interaction, where the meaning of the current emotional expression depends on a prior trajectory rather than on the current signal alone. The second is multimodal conflict, where one modality is noisy or misleading and historical affective memory is needed to calibrate interpretation. The third is modality incompleteness, where memory and the remaining available channels must work together to preserve continuity of judgment.

In this report, these ideas are operationalized through the Memory Bear AI Memory Science Engine, which integrates structured emotional memory formation, retrieval, dynamic fusion calibration, and continuous memory updating into a unified architecture.

\subsection{Contributions of This Technical Report}

This technical report presents the \textbf{Memory Bear AI Memory Science Engine} as a system-level framework for multimodal affective intelligence. Its contributions can be summarized as follows.

\begin{itemize}
    \item \textbf{A memory-centered perspective on affective judgment.}  
    We formulate multimodal affective understanding as a problem that requires not only perception and fusion, but also the organized preservation, retrieval, and updating of affective information across interaction horizons\cite{maharana2024locomo,jia2025longtermmemory}.

    \item \textbf{A structured architecture for memory-driven affective processing.}  
    We present an architecture that integrates multimodal preprocessing, structured emotional memory formation, working-memory aggregation, long-term consolidation, associative retrieval, dynamic fusion modulation, and memory lifecycle management into a unified pipeline.

    \item \textbf{A memory-guided mechanism for robust multimodal interpretation.}  
    We introduce a fusion strategy in which multimodal contributions are modulated not only by current signal reliability, but also by their consistency with historically relevant affective memory.

    \item \textbf{A technical report perspective connecting architecture, robustness, and deployment relevance.}  
    Beyond describing the model structure itself, this report discusses why memory-centered affective judgment is particularly valuable in realistic settings involving long interaction horizons, missing modalities, and unstable signal quality.
\end{itemize}

The remainder of this report is organized as follows. Chapter 2 reviews the technical background and identifies major gaps in existing multimodal affective systems. Chapter 3 presents the design philosophy of the Memory Bear AI Memory Science Engine. Chapter 4 describes the architecture of the memory-driven affective engine in detail. Chapter 5 discusses experimental validation and case-based analysis. Chapters 6 and 7 examine system advantages, limitations, and future directions, followed by the conclusion in Chapter 8.

\section{Background and Technical Gaps}

\subsection{Multimodal Affective Modeling Beyond Local Perception}

Multimodal affective modeling has developed rapidly because human emotion is rarely expressed through a single channel. Language, speech, and visual behavior often provide complementary evidence, and recent multimodal systems have significantly improved affective prediction by modeling these signals jointly \cite{lian2024merbench,ramaswamy2024merreview,arunagladys2023survey}. Compared with unimodal approaches, multimodal methods are better able to capture weak, indirect, and cross-channel emotional cues, especially when one modality alone is insufficient for reliable interpretation.

\begin{figure}[t]
    \centering
    \includegraphics[width=0.8\linewidth]{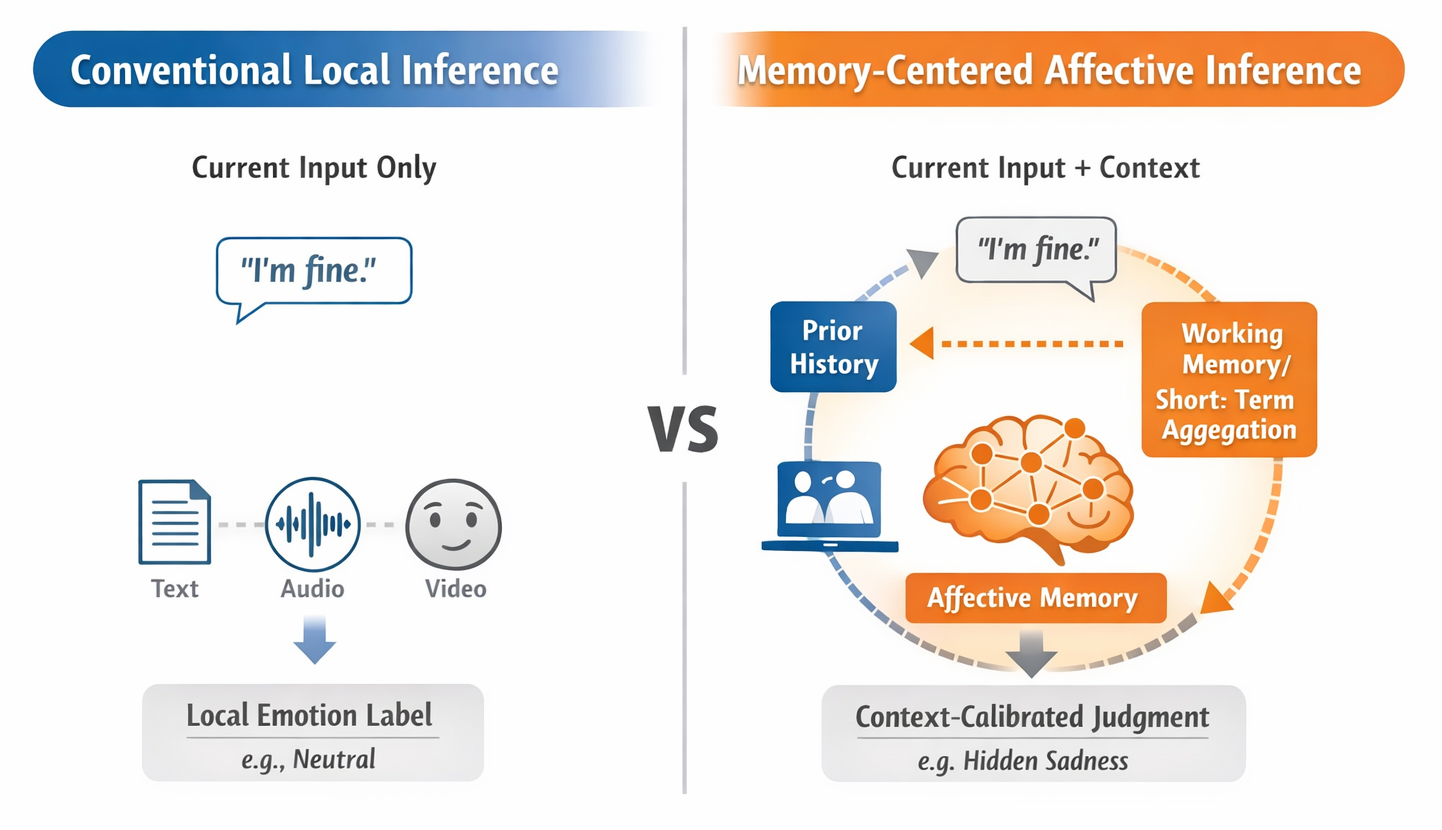}
    \caption{Motivational comparison between conventional local affective inference and memory-centered affective inference. Conventional multimodal systems typically rely on current-turn text, audio, and visual cues to produce a local emotion label, whereas memory-centered inference integrates current input with prior interaction history and affective memory to produce a context-calibrated judgment.}
    \label{fig:problem_motivation}
\end{figure}

This progress is important, but it does not fully resolve the problem addressed in this report. Most multimodal affective systems still operate primarily within a local perceptual window. Even when stronger fusion mechanisms are used, the main objective remains the inference of the current emotional state from currently available multimodal evidence. In other words, multimodal modeling has become more expressive, but it is still often centered on present-state prediction rather than persistent affective judgment.

This distinction matters because real affective interpretation frequently depends on more than concurrent multimodal input. Emotional meaning may be weakly expressed in the current turn, shaped by prior interaction, or only recoverable when current signals are interpreted against a broader affective trajectory. As a result, stronger multimodal fusion alone does not necessarily provide continuity, reinterpretation, or robustness when present evidence is incomplete, noisy, or emotionally indirect.


\subsection{Memory-Related Approaches in Affective Modeling}

The limitations of purely local perception have motivated growing interest in memory-related approaches to affective modeling. Although this line of work does not follow a single unified taxonomy, existing methods broadly suggest that affective interpretation benefits from some form of contextual retention beyond the current local input \cite{shou2023mcer_survey,hazarika2018cmn,wen2023dimmn}.

A first line of work uses \emph{implicit sequential memory}, typically through recurrent architectures such as LSTMs or GRUs. These models preserve short-range context in hidden states and are useful for capturing local temporal continuity. A second line of work introduces \emph{explicit contextual memory}, such as memory-network-based or conversational memory approaches, in which prior context can be stored and later queried during inference \cite{hazarika2018cmn}. A third line of work explores \emph{memory-supported multimodal interaction}, where memory is used not only to retain history but also to mediate cross-modal interpretation and fusion \cite{wen2023dimmn}. Together, these directions show that memory has already become an important idea in affective modeling, especially in conversational and multimodal settings.

For the purpose of this report, however, the key issue is not to provide a fine-grained taxonomy of prior memory methods. The more important observation is that most existing approaches still use memory in relatively limited ways: as hidden-state carryover, contextual lookup, or local fusion support. They do not yet fully treat affective information as a structured memory object that can be explicitly encoded, prioritized, retrieved, revised, and selectively forgotten across interaction horizons.

A particularly relevant neighboring direction concerns \emph{robust multimodal learning under missing or degraded input}. Recent surveys on missing-modality learning emphasize that incomplete, corrupted, or asynchronous multimodal evidence is common in real-world systems \cite{wu2024missingmodalitysurvey}. This point is especially important for the present report, because many practical affective scenarios involve exactly these conditions: uneven modality availability, unstable signal quality, and emotionally ambiguous local input. From this perspective, memory is valuable not only for retaining history, but also for stabilizing interpretation when current multimodal evidence is weak or partially unavailable.
This limitation becomes central to the Memory Bear AI design introduced next.

\subsection{Technical Gaps in Current Affective Systems}

Despite the progress outlined above, current multimodal affective systems still exhibit several gaps that limit their ability to support persistent and context-sensitive affective judgment.

A first gap is the continued dominance of \emph{snapshot-oriented inference}. Even advanced multimodal systems are often designed to infer emotion from the current utterance, segment, or short interaction window. This works well for benchmark-style prediction, but it remains limited when emotional meaning depends on accumulated interaction history, prior unresolved context, or gradual affective drift over time \cite{lian2024merbench,ramaswamy2024merreview,arunagladys2023survey}.

A second gap is the \emph{lack of structured affective memory}. In many systems, multimodal evidence is processed and fused for immediate prediction, but it is not preserved as an explicit memory object that can later support retrieval and reinterpretation. Historically meaningful affective information therefore remains dispersed across latent states or modality-specific features instead of being organized into reusable memory representations.

A third gap is the \emph{limited lifecycle management of emotional information}. Existing memory-related approaches often preserve context or support retrieval, but they less frequently distinguish among short-term aggregation, selective consolidation, retrieval priority, revision, and adaptive forgetting. As a result, memory may be present, but it is not fully managed.

A fourth gap is the \emph{underdeveloped treatment of emotion as a memory variable}. Prior work has demonstrated the usefulness of context and memory in affective modeling \cite{hazarika2018cmn,shou2023mcer_survey,wen2023dimmn}, but emotion is still often treated mainly as a prediction target rather than as something that should be encoded together with reliability, context, salience, and temporal structure.

A fifth gap is the \emph{separation between robustness and memory}. Under noisy, conflicting, or incomplete multimodal conditions, many systems still depend primarily on local compensation or degraded inference. They do not fully exploit historical affective context as a stabilizing factor when current evidence is weak or partially unavailable \cite{pham2026missbench,wu2024missingmodalitysurvey,xue2025mmrc}.

The next section outlines how the Memory Bear AI design responds to these limitations.

\subsection{Design Motivation of the Memory Bear AI Engine}

The Memory Bear AI framework is motivated by exactly this gap between local multimodal perception and persistent affective judgment. In this report, \emph{memory science} refers to a system design perspective in which affective
information is not only perceived, but also encoded into reusable memory representations,
selectively retained, reactivated when contextually relevant, and revised as new evidence
accumulates. This perspective aligns with the broader Memory Bear framework, which
treats memory as a core infrastructure for overcoming the limitations of short-context
LLM systems and for supporting more adaptive cognitive services \cite{wen2025memorybear}.

From this perspective, the problem is not whether multimodal affective modeling should use stronger encoders or stronger fusion alone. The deeper question is how affective meaning can persist beyond the current moment and later influence interpretation when present evidence is weak, incomplete, or ambiguous. This is where the Memory Bear AI design departs from conventional approaches.

The proposed framework therefore treats memory as the organizing layer of affective judgment. Multimodal affective evidence is transformed into structured emotional memory representations that preserve emotional semantics together with contextual, source-related, and temporal attributes. These representations are later formalized as Emotion Memory Units (EMUs) in the architecture chapter and serve as the basis for short-term aggregation, long-term consolidation, retrieval, memory-guided fusion, and continuous memory updating.

The Memory Bear AI engine is therefore introduced as a memory-centered affective architecture rather than as a stronger local classifier with an auxiliary memory module. The next chapter develops this idea at the design level, followed by the system architecture.

\section{Design Philosophy of the Memory Bear AI Memory Science Engine}

\subsection{Memory as Cognitive Infrastructure}

A central premise of the Memory Bear AI framework is that memory should not be treated
as a passive store of historical data. Instead, it functions as a core cognitive infrastructure
that supports continuity, retrieval, adaptation, and reasoning across interaction horizons,
an idea that is also emphasized in the broader Memory Bear system design
\cite{wen2025memorybear}.

This perspective is particularly important in multimodal settings. Text, speech, and visual behavior often provide partial or conflicting affective evidence. A local fusion system can only interpret these signals within the current perceptual window, whereas a memory-centered system can compare current evidence against prior affective traces and use that comparison to stabilize interpretation. Under this view, memory functions as the organizing substrate through which multimodal affective signals acquire continuity, relevance, and interpretive value.

Within the Memory Bear AI framework, memory serves four system-level functions. It organizes affective evidence into reusable representations, preserves continuity beyond the local processing window, supports contextual calibration under noisy or incomplete conditions, and enables adaptive evolution through consolidation, retrieval, updating, and selective forgetting. For this reason, the Memory Bear AI engine is built around a memory substrate that transforms multimodal affective evidence into structured, reusable, and evolvable cognitive objects.

\subsection{Emotional Memory as a Native Cognitive Dimension}

A second principle of the Memory Bear AI framework is that emotion should not be treated merely as an output label appended to the end of inference. Instead, emotion is modeled as a \emph{native cognitive dimension} of memory itself. This perspective is supported by a broad body of cognitive and affective science showing that emotion shapes how information is attended to, encoded, consolidated, retrieved, and interpreted over time \cite{tyng2017emotionmemory,labar2007emotionalmemory,faul2022moodcongruent}.

From a cognitive perspective, emotional memory is not equivalent to factual storage. Emotional significance influences what is prioritized during encoding, which components of an episode are retained with greater strength, and how later retrieval is biased by current or prior affective context \cite{kensinger2009details,faul2022moodcongruent}. Emotion is therefore not only something that is remembered; it is also part of the mechanism that determines how remembering occurs.

This point is especially relevant for interactive AI systems. If emotion is treated only as a terminal prediction target, then once an emotional label has been produced, its role in later interpretation remains weak. By contrast, if emotion is treated as a native dimension of memory organization, then affective information can shape retrieval priority, contextual calibration, and decision support in subsequent interaction. Emotional traces are therefore not stored as detached labels, but as structured memory objects linked to source reliability, contextual anchors, intensity, and temporal position.

A useful example is that the same observable event can be interpreted differently under different emotional histories. Consider a short user message such as ``I guess that's fine.'' In a local system, this utterance may be classified as neutral or mildly positive. However, if it appears after repeated setbacks and unresolved tension, it may instead signal resignation or suppressed frustration. Conversely, if it follows successful resolution of earlier difficulty, the same utterance may be more plausibly interpreted as relief or acceptance. The wording itself remains nearly unchanged, yet its emotional meaning shifts because the relevant memory background is different \cite{faul2022moodcongruent,clore2007judgment}.

Accordingly, the Memory Bear AI Memory Science Engine adopts a representation strategy in which emotion is embedded directly into the structure of memory. These structured representations are later formalized as Emotion Memory Units (EMUs) in the architecture section.

\subsection{Three Core Principles of the Engine}

Building upon the view of memory as cognitive infrastructure and emotion as a native cognitive dimension, the Memory Bear AI framework can be summarized through three core design principles.

\paragraph{Principle 1: Emotional understanding must be history-aware.}
Emotional meaning cannot be reliably inferred from the present moment alone. In natural interaction, affective states are often shaped by accumulated experience, unresolved prior events, repeated activation, interpersonal history, and longer-term contextual dynamics. For the Memory Bear AI engine, this means that affective processing must be grounded in access to historically relevant emotional traces rather than restricted to short-window inference \cite{maharana2024locomo,tyng2017emotionmemory,faul2022moodcongruent}.

\paragraph{Principle 2: Present multimodal interpretation should be memory-calibrated.}
History is not only something that should be available; it should actively modulate how current multimodal evidence is interpreted. In the proposed framework, textual, acoustic, and visual cues are evaluated not only by their immediate reliability, but also by their consistency with historically relevant affective memory. A weak signal may become highly informative when it aligns with prior context, whereas a salient signal may deserve lower interpretive weight if it conflicts with a more stable emotional trajectory. It is also consistent with recent work showing that memory-aware and context-aware conversational models can improve emotion recognition, although they typically stop short of modeling a full affective memory lifecycle \cite{hazarika2018cmn,wen2023dimmn,shou2023mcer_survey}.

\paragraph{Principle 3: Affective memory must be selectively consolidated, prioritized, and forgotten.}
Affective memory is valuable because it evolves selectively rather than accumulating indiscriminately. Emotionally intense, highly activated, or decision-relevant traces should receive higher consolidation priority, stronger retention strength, and greater retrieval salience, whereas low-value or outdated traces should be gradually deprioritized, merged, revised, or forgotten \cite{labar2007emotionalmemory,kensinger2009details,tyng2017emotionmemory}. This principle implies that affective memory must be managed as a lifecycle rather than as a static archive.

Together, these principles define the shift from short-term recognition to persistent multimodal affective intelligence and provide the conceptual basis for the architecture described next.

\subsection{From Snapshot Perception to Persistent Affective Understanding}

The design philosophy of the Memory Bear AI framework can be understood as a transition from \emph{snapshot perception} to \emph{persistent affective understanding}. In conventional multimodal emotion recognition systems, the dominant objective is to infer the emotional state expressed in the current utterance, segment, or short interaction window. Even when contextual modeling is included, the overall processing logic often remains centered on present-state prediction.

\begin{figure}[H]
    \centering
    \includegraphics[width=\linewidth]{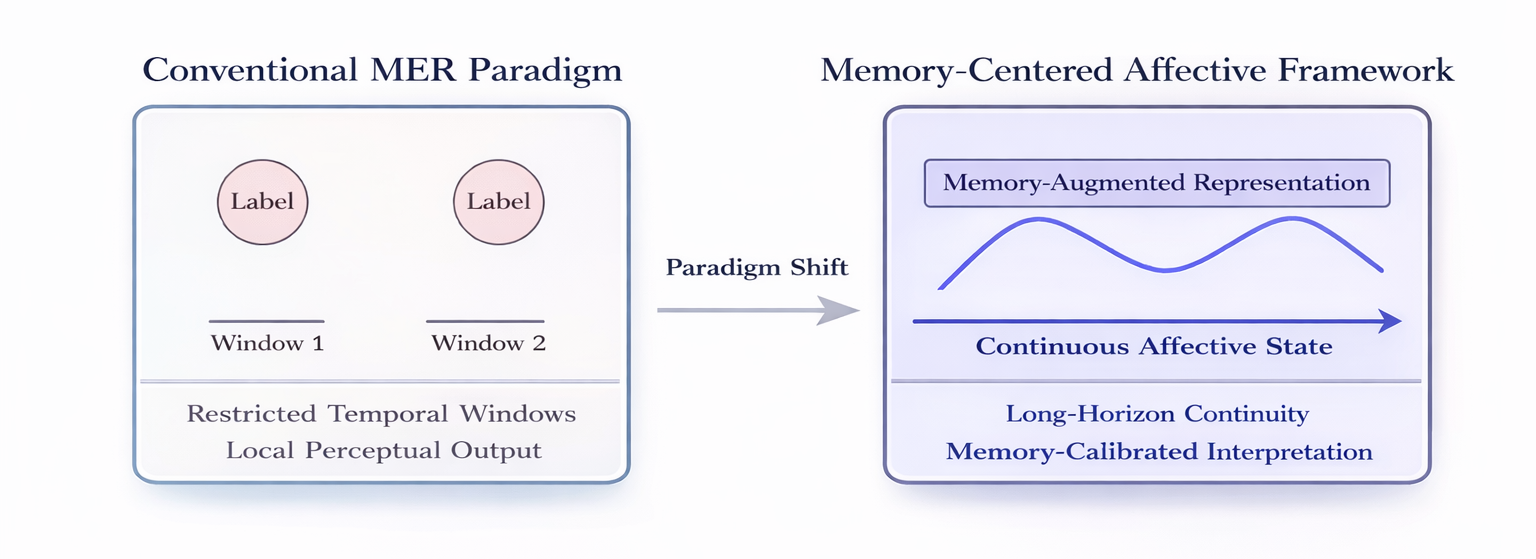}
    \caption{Conceptual shift from snapshot-based multimodal emotion recognition to a memory-centered framework for persistent affective understanding.}
    \label{fig:snapshot2memorycentered}
\end{figure}

The Memory Bear AI perspective adopts a different assumption: each new affective observation should be interpreted in relation to an evolving memory structure. Current multimodal evidence is therefore evaluated not in isolation, but against historically relevant affective memory and recent short-term affective state. Under this formulation, affective understanding becomes cumulative rather than episodic.

This shift changes the role of memory, emotion, and fusion simultaneously. Memory becomes the substrate through which affective signals acquire continuity. Emotion becomes part of memory organization rather than only a prediction target. Multimodal fusion becomes historically calibrated interpretation rather than local signal combination alone.

For interactive AI systems, this distinction is consequential. Sustained emotional support, companion dialogue, customer interaction, education, and health-related applications all require more than momentary affect recognition. They require systems that can maintain continuity, recover relevant prior context, and interpret new emotional evidence in light of what has been previously experienced.

This transition from snapshot perception to persistent affective understanding provides the conceptual basis for the architecture presented in the next chapter.

\section{Architecture of the Memory Bear AI Engine}

\subsection{Overall Framework}

To operationalize the memory-centered perspective developed in the preceding chapters, the Memory Bear AI Memory Science Engine adopts a four-stage architecture for multimodal affective processing. The framework is designed to transform heterogeneous affective evidence from text, speech, and vision into a persistent and structured form of emotional understanding.

\begin{figure}[t]
    \centering
    \includegraphics[width=\linewidth]{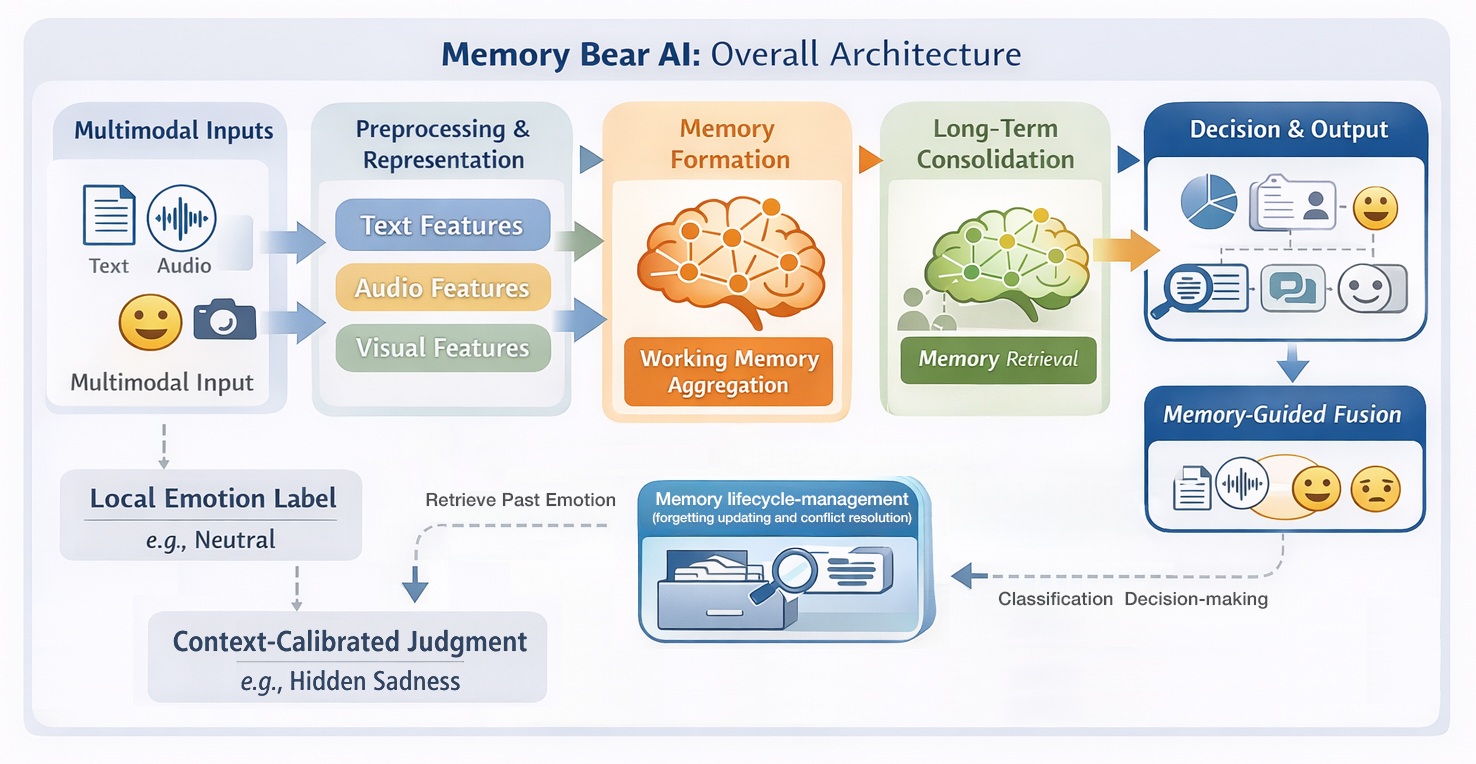}
    \caption{Overall architecture of the Memory Bear AI framework. The system processes multimodal inputs through preprocessing, memory formation, consolidation, retrieval, memory-guided fusion, and affective decision-making, while continuously updating memory based on new interaction evidence.}
    \label{fig:overall_architecture}
\end{figure}

\begin{figure}[H]
    \centering
    \includegraphics[width=0.6\linewidth]{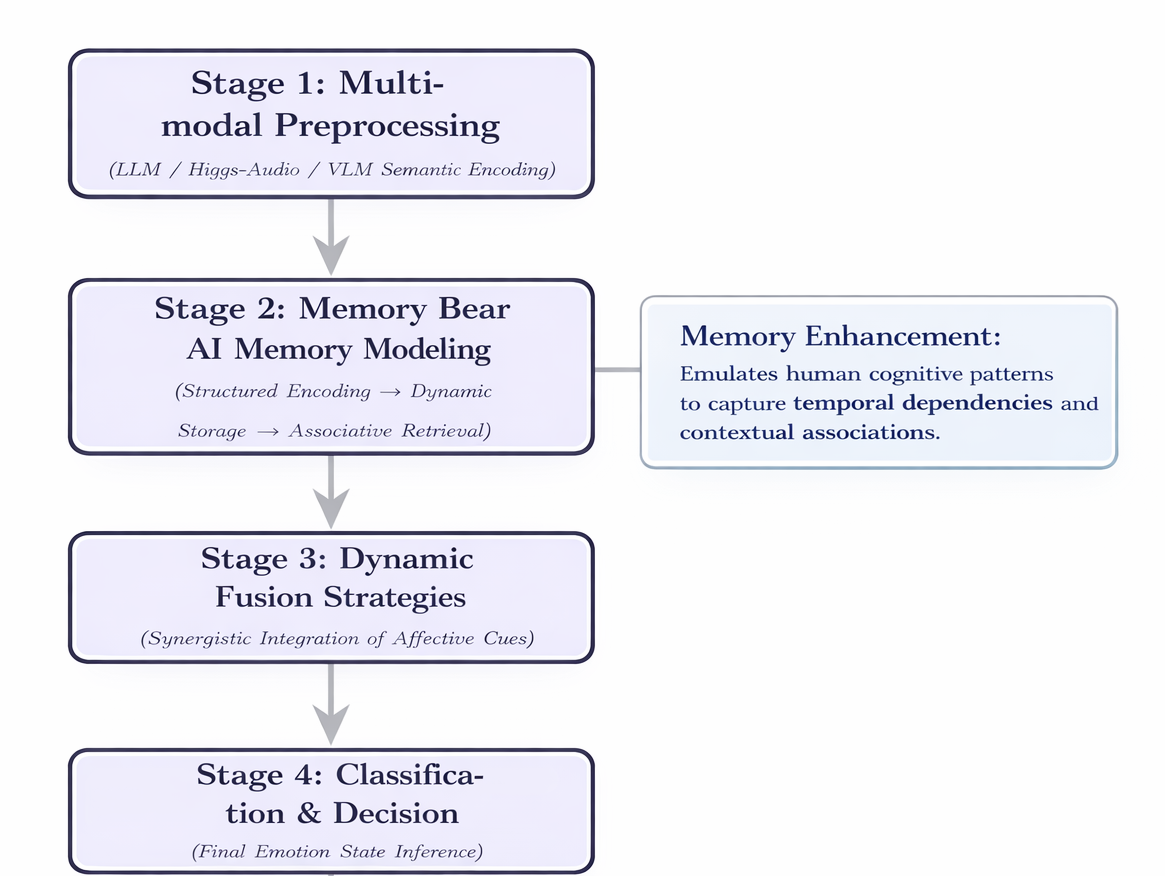}
    \caption{The four-stage architecture of the memory-driven affective engine.}
    \label{fig:four_stage_arch}
\end{figure}

As illustrated in Figure~\ref{fig:four_stage_arch}, the architecture consists of four major stages: \textbf{(1) multimodal preprocessing and representation learning}, \textbf{(2) Memory Bear AI memory modeling}, \textbf{(3) dynamic fusion strategies}, and \textbf{(4) classification, decision-making, and memory updating}. These stages correspond, respectively, to multimodal perception, structured affective memory formation, memory-guided interpretation, and post-inference affective decision with memory evolution.

The overall design follows the core logic of the Memory Bear AI framework: affective information should first be encoded into semantically meaningful modality-specific representations, then organized into structured memory, used to calibrate the interpretation of current multimodal evidence, and finally incorporated into an evolving memory system through decision-dependent updating. In this way, the engine forms a closed loop linking perception, memory, fusion, decision, and memory management.

\subsection{Advanced Multimodal Representation Learning}

The first stage of the Memory Bear AI engine focuses on multimodal preprocessing and representation learning. Its main purpose is to convert raw text, speech, and visual input into high-level affect-relevant representations that can support the subsequent memory layer. At this stage, the goal is not yet structured memory formation, but the preparation of semantically meaningful modality-specific representations for later encoding, retrieval, and fusion.

The framework operates over three primary modalities: text, speech, and vision. Each modality contributes complementary affective evidence. Text provides semantic content and discourse-level emotional cues. Speech captures prosody, rhythm, pitch variation, timbre, and other paralinguistic information. Vision contributes facial expression, movement, posture, and broader nonverbal behavioral signals. To preserve these different aspects of affective information, the present framework adopts dedicated encoders for each modality.

\begin{figure}[H]
    \centering
    \includegraphics[width=\linewidth]{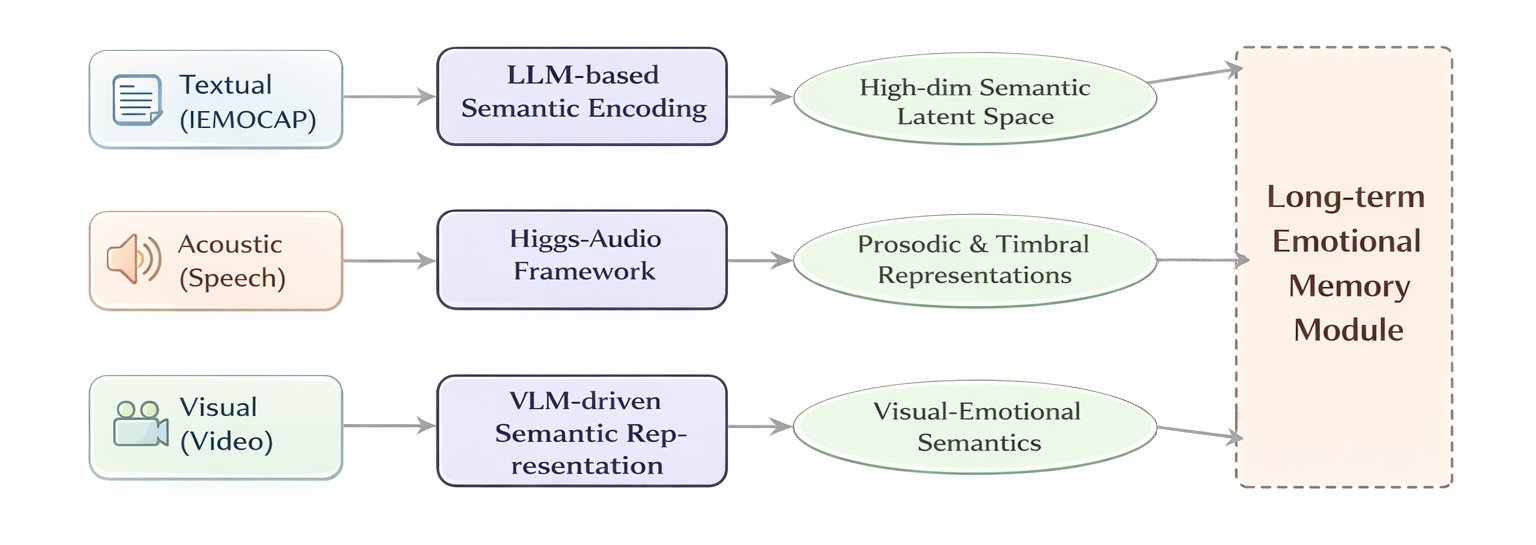}
    \caption{Detailed architecture of the multimodal representation learning pipeline.}
    \label{fig:pipeline_design}
\end{figure}

\paragraph{Textual modality: LLM-based semantic encoding.}
For textual input, the framework employs a large-language-model-based encoder to obtain contextualized semantic representations. This choice supports the interpretation of indirect, implicit, or context-dependent emotional meaning in language, including suppression, irony, mitigation, and discourse-sensitive affective cues.

\paragraph{Acoustic modality: Higgs-Audio-based representation.}
For speech input, the framework adopts Higgs-Audio as the primary acoustic representation module. Higgs-Audio is used to capture affect-relevant vocal patterns such as pitch dynamics, vocal tension, energy variation, rhythm, and pause structure, providing a richer acoustic basis for subsequent emotion-related memory modeling.

\paragraph{Visual modality: VLM-driven semantic representation.}
For visual input, the framework employs a Vision-Language-Model-driven encoder to obtain semantically oriented visual representations. Compared with purely low-level handcrafted descriptors, this design better preserves higher-level visual-affective information, including expression patterns, head movement, posture, and context-sensitive nonverbal cues.

Overall, Stage 1 serves as the perceptual front-end of the engine. Its output is not yet a memory representation, but a set of modality-specific affective encodings that prepare the input for the Memory Bear AI memory modeling stage described next.

\subsection{Structured Affective Memory Modeling}

The second stage of the Memory Bear AI engine is \emph{structured affective memory modeling}, which forms the structural core of the proposed system. If Stage 1 provides semantically enriched multimodal perceptual representations, Stage 2 transforms those representations into structured affective memory that can be preserved, organized, retrieved, and reused over time. The purpose of the memory layer is to convert transient multimodal evidence into a form that supports continuity-aware affective reasoning.

This stage is motivated by a central limitation of many conventional multimodal pipelines: even when they capture rich local affective features, they often lack an explicit representational layer in which emotional information can be stored as reusable memory objects. As a result, historically meaningful affective cues may remain dispersed across latent states or modality-specific features, making later retrieval, calibration, and updating difficult to perform in a principled way. The Memory Bear AI framework addresses this limitation by organizing multimodal affective evidence into structured memory units and by explicitly distinguishing between short-term affective aggregation, longer-term retention, and memory-guided retrieval.

Concretely, the memory modeling stage is composed of four closely related components: \textbf{(1) multimodal emotion encoding and structured memory formation}, \textbf{(2) emotion working memory and short-term aggregation}, \textbf{(3) emotion long-term memory and consolidation}, and \textbf{(4) memory-driven retrieval}. The following subsections describe these components in sequence, beginning with the Emotion Memory Unit (EMU) as the basic structured object of the memory layer.

\subsubsection{Multimodal Emotion Encoding and Structured Memory Formation}

The multimodal emotion encoding module is the entry point of the Memory Bear AI memory layer. Its purpose is to transform heterogeneous affective signals from text, speech, and vision into a unified and structured representation that can serve as the basic object of memory operations. Rather than directly passing multimodal embeddings into downstream fusion or classification, the present framework first converts them into explicit memory units so that emotional evidence can be stored, retrieved, reprioritized, and updated over time.

This design addresses a common weakness in conventional multimodal emotion pipelines. In many systems, multimodal information is combined for immediate prediction, but the resulting representation does not explicitly preserve emotional semantics, source reliability, contextual relevance, intensity, and temporal position as organized memory attributes. This makes later retrieval, prioritization, and reinterpretation more difficult.

To overcome this limitation, the Memory Bear AI framework introduces the \emph{Emotion Memory Unit (EMU)} as the basic representational object of affective memory. Each EMU is intended to capture a moment of multimodal affective evidence in a structured form that can later participate in aggregation, consolidation, retrieval, and updating. The EMU is defined as:
\[
\mathrm{EMU}_t = \{ e_t,\; m_t,\; c_t,\; \alpha_t,\; \tau_t \}
\]
where \(e_t\) denotes the emotion category or continuous affective vector, \(m_t\) denotes modality source and reliability, \(c_t\) denotes the contextual semantic anchor, \(\alpha_t\) denotes affective intensity or salience weight, and \(\tau_t\) denotes temporal information.

\begin{figure}[t]
    \centering
    \includegraphics[width=\linewidth]{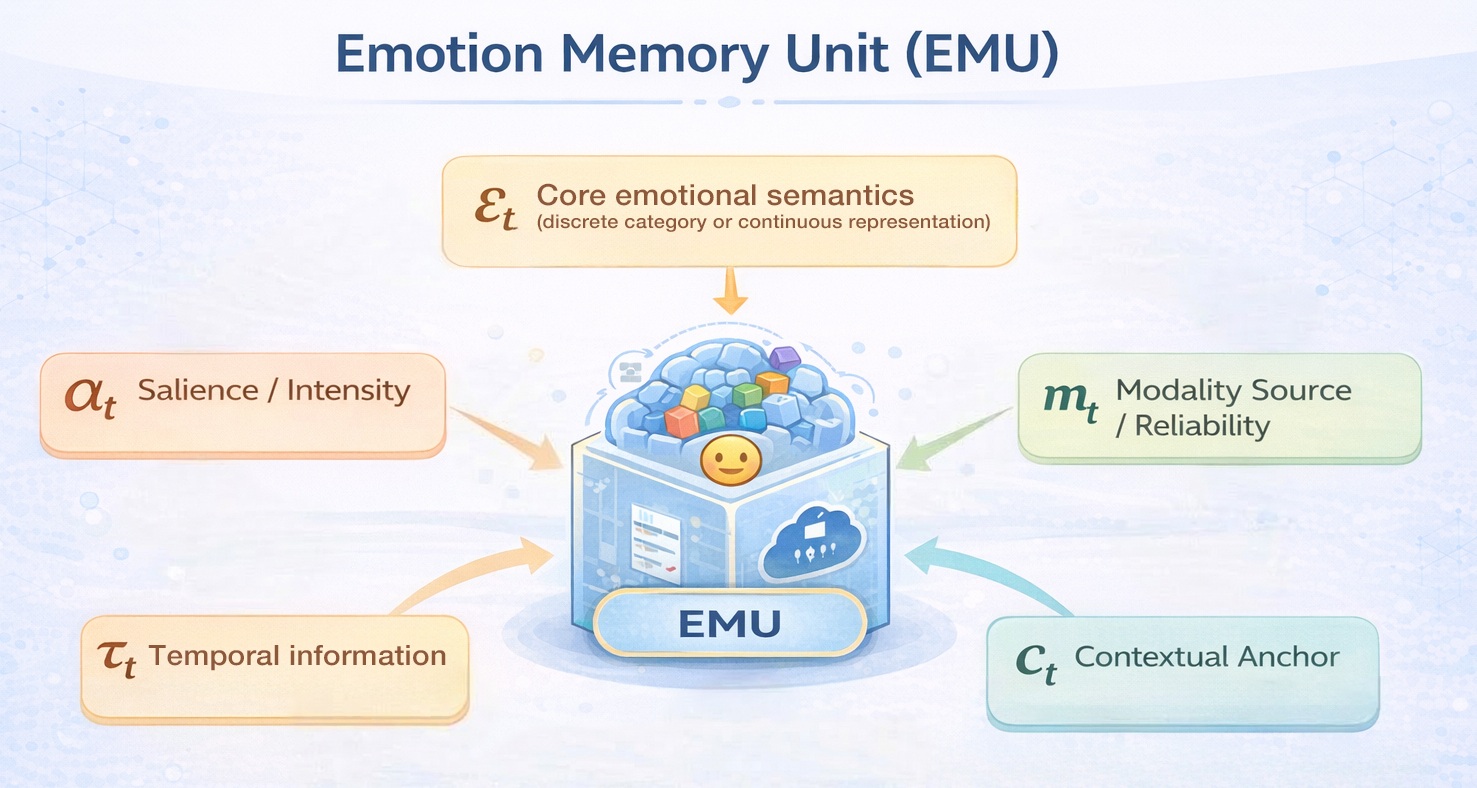}
    \caption{Structured representation of the Emotion Memory Unit (EMU). Each EMU encodes affective semantics together with source-related reliability, contextual anchoring, salience, and temporal information.}
    \label{fig:emu_structure}
\end{figure}

The first dimension, \(e_t\), represents the core emotional semantics of the current observation. Depending on the task formulation, this component may take the form of a discrete emotion category or a continuous affective representation such as a valence-arousal vector. Its role is to provide the primary affective meaning associated with the current multimodal input.

The second dimension, \(m_t\), records modality source and reliability. This component is crucial because different modalities often contribute unequally under real-world conditions. Text may provide strong semantic cues while speech is noisy; acoustic cues may remain informative when language is emotionally restrained; visual evidence may be unreliable because of occlusion or weak signal quality. By explicitly encoding source-related reliability, the memory layer preserves information that can later support memory-guided fusion and robustness-oriented interpretation.

The third dimension, \(c_t\), represents the contextual semantic anchor. This component links the current emotional observation to the situational, conversational, or semantic context in which it arises. The contextual anchor is especially important for associative retrieval, because later memory access should not depend only on abstract emotional similarity, but also on whether the current situation is meaningfully related to prior affective experience.

The fourth dimension, \(\alpha_t\), denotes affective intensity or salience. In the Memory Bear AI framework, not all emotional traces are treated equally. Emotionally intense, highly activated, or decision-relevant observations should receive higher retention priority and greater retrieval salience than weak or low-value traces. The salience weight therefore provides a direct connection between emotional intensity and later memory management, including consolidation strength, retrieval priority, and resistance to forgetting.

The fifth dimension, \(\tau_t\), records temporal information. Since the present framework is explicitly concerned with long-horizon affective continuity, temporal position is not an auxiliary detail but part of the core structure of emotional memory. Temporal tags allow the system to distinguish between recent and distant affective traces, preserve chronological ordering, and later support recency-sensitive retrieval and adaptive forgetting.

The EMU serves as the basic representational unit of the memory layer and provides the input to subsequent working-memory aggregation and long-term consolidation.

\subsubsection{Emotion Working Memory and Short-Term Aggregation}

Once multimodal affective evidence has been encoded into Emotion Memory Units (EMUs), the resulting memory objects first enter the \emph{Emotion Working Memory} (E-WM) layer. This component provides the short-horizon processing space of the Memory Bear AI engine, supporting the immediate interpretation of affective signals within the current interaction window. In contrast to long-term memory, which is responsible for persistent retention and later retrieval, working memory focuses on the temporary organization of recent emotional observations before selective consolidation takes place.

The introduction of an explicit working-memory layer is motivated by an important property of affective interaction: emotional meaning often does not emerge from a single isolated observation, but from a short sequence of related cues unfolding over nearby moments. A brief pause, a slight change in vocal tension, and a subtle shift in wording may each be weak in isolation, yet collectively indicate a meaningful transition in affective state. If such cues are processed only as independent local observations, transient noise may be overemphasized and short-range emotional evolution may be missed. The role of working memory is therefore to provide a structured intermediate layer in which recent EMUs can be aggregated into a more stable short-term affective representation.

Formally, let the working-memory state at time \(t\) be denoted by:
\[
E^{\mathrm{STM}}_t = \sum_{i=t-k+1}^{t} w_i \cdot \mathrm{EMU}_i
\]
where \(k\) denotes the size of the current working-memory window, \(w_i\) denotes the weight assigned to the emotional memory unit at time \(i\), and \(\mathrm{EMU}_i\) denotes the structured affective memory object generated from multimodal evidence at that moment. The resulting representation \(E^{\mathrm{STM}}_t\) can be understood as the short-term affective state maintained within the current interaction horizon.

This formulation reflects two important design choices. First, short-term affective understanding is treated as an aggregation process rather than as a single-step decision based only on the latest input. Second, recent affective observations do not contribute equally: their influence can vary as a function of recency, salience, contextual relevance, or modality reliability. In practice, this allows the working-memory layer to smooth isolated fluctuations while remaining sensitive to genuine local changes in emotional trajectory.

The working-memory layer serves several functions in the overall system. It reduces the influence of transient noise by integrating nearby affective evidence into a more coherent short-term state. It preserves short-range emotional evolution, allowing the system to represent local shifts such as escalation, softening, hesitation, or recovery. It also provides a higher-value candidate representation for later long-term consolidation, since not every instantaneous EMU should be written directly into long-term memory without short-term filtering and aggregation.

Within the Memory Bear AI framework, working memory serves as an intermediate layer between immediate multimodal perception and persistent affective retention. It captures what is currently active, emotionally relevant, and locally coherent, while also preparing the system to decide which affective traces should later be strengthened, retained, or discarded.

The output of this layer is not yet a persistent memory record. Instead, it provides the short-term affective substrate on which the subsequent long-term memory consolidation mechanism operates.

\subsubsection{Emotion Long-Term Memory and Consolidation Mechanism}

While the working-memory layer maintains a short-horizon affective state for ongoing interpretation, persistent affective intelligence requires a second layer in which selected emotional information can be retained over longer interaction horizons. For this purpose, the Memory Bear AI engine introduces an \emph{Emotion Long-Term Memory} (E-LTM) mechanism, whose role is to preserve affective traces that are sufficiently salient, recurrent, or decision-relevant to remain useful beyond the current interaction window.

The introduction of long-term affective memory is motivated by a basic asymmetry in emotional interaction. Not all observed emotional cues deserve persistent retention. Many are transient, weak, or highly local, and should remain confined to short-term processing. Others, however, carry broader interpretive value. Repeated frustration, persistent hesitation, recurring reassurance, prolonged emotional drift, and high-intensity affective episodes may all shape the interpretation of future interaction. A system that lacks long-term consolidation treats these patterns as disposable local observations, whereas a memory-centered system should preserve them as part of an evolving affective history.

In the Memory Bear AI framework, long-term memory formation is governed by a \emph{selective consolidation} process. Rather than writing every short-term affective state directly into persistent storage, the engine determines whether recent emotional evidence should be consolidated based on factors such as affective salience, repeated activation, contextual importance, and decision relevance. This design follows the principle established earlier in the report: affective memory must be selectively consolidated, prioritized, and forgotten rather than accumulated indiscriminately.

Let the short-term affective state at time \(t\) be denoted by \(E^{\mathrm{STM}}_t\). The consolidation process determines whether this state, or part of it, should be incorporated into long-term memory:
\[
\mathcal{M}^{\mathrm{LTM}}_{t+1} = \mathcal{M}^{\mathrm{LTM}}_{t} \cup \mathrm{Consolidate}(E^{\mathrm{STM}}_t)
\]
where \(\mathcal{M}^{\mathrm{LTM}}_{t}\) denotes the current long-term affective memory store and \(\mathrm{Consolidate}(\cdot)\) denotes the selective operation that decides whether the present short-term affective state should be transformed into a persistent memory element. This formulation is intentionally abstract at the architectural level: the key point is not that all short-term states are retained, but that consolidation is conditional and memory-aware.

Consolidation serves at least three important functions. First, it filters local affective evidence, preventing the long-term memory from being flooded by transient or low-value observations. Second, it preserves emotionally significant patterns that may later become essential for contextual calibration and retrieval. Third, it creates continuity across interaction horizons by allowing affective meaning to persist even after the local perceptual context has passed.

Affective salience is especially important in this process. Emotionally intense, highly activated, or repeatedly re-encountered traces should receive stronger consolidation priority than weak or incidental observations. Likewise, affective states that substantially influence decision-making or subsequent interaction should be retained with higher importance. In this way, long-term consolidation is not only a retention mechanism, but also an early stage of memory prioritization: it determines which emotional experiences become part of the system's persistent affective substrate.

The resulting E-LTM is not organized as a flat archive. Instead, it stores structured affective memory objects that preserve emotional semantics together with contextual and temporal attributes, thereby supporting later retrieval and reinterpretation. Once consolidated, these memory objects can contribute to future affective processing even when the original interaction moment is no longer present in the local perceptual window.

E-LTM provides the basis for affective continuity. It allows the engine to carry forward historically meaningful emotional traces beyond the limits of immediate perception and short-term aggregation.

\subsubsection{Memory-Driven Retrieval Mechanism}

The value of long-term affective memory does not lie in retention alone. For persistent affective intelligence, historically stored emotional information must remain accessible to later reasoning and capable of influencing present interpretation when contextually relevant. For this reason, the final component of the Memory Bear AI memory modeling stage is a \emph{memory-driven retrieval mechanism}, through which the system reactivates historically relevant emotional traces from Emotion Long-Term Memory (E-LTM) and makes them available for current affective inference.

This mechanism addresses a central limitation of many conventional multimodal systems. Even when prior context is encoded somewhere in the model, historically meaningful affective information often remains entangled in hidden states or distributed latent representations, making it difficult to access in a targeted and interpretable manner. In contrast, the Memory Bear AI framework treats retrieval as an explicit operation over structured emotional memory. The goal is to identify memory traces that are contextually relevant to the present interaction rather than merely emotionally similar to it.

Formally, let the current contextually grounded query be denoted by \(q_t\), which may be constructed from the present multimodal observation together with the short-term affective state. The retrieval process can then be written as:
\[
M_t = \mathrm{Retrieve}(\mathcal{M}^{\mathrm{LTM}}_t, q_t)
\]
where \(\mathcal{M}^{\mathrm{LTM}}_t\) denotes the current long-term affective memory store and \(M_t\) denotes the subset of historically relevant emotional memory returned for time \(t\). At the architectural level, the specific retrieval function may be implemented through similarity matching, context-conditioned attention, or another memory-association mechanism. The essential point is that retrieval is guided by present contextual relevance rather than by indiscriminate access to the entire memory store.

The retrieval query is not based solely on the current emotional label. Instead, it may draw upon several memory-relevant cues already encoded in the system, including current affective semantics, contextual anchors, recent short-term affective state, and source-related characteristics of the present input. This design is important because affective retrieval should not depend only on coarse emotional similarity. Two emotionally similar states may still differ in conversational role, situational cause, interpersonal meaning, or temporal significance. By incorporating contextual anchors into the retrieval process, the system is better able to reactivate the affective history that is actually relevant to the present interaction.

Memory-driven retrieval serves at least three functions. First, it restores historically meaningful affective context that is no longer present in the local perceptual window. Second, it enables current multimodal interpretation to be calibrated by prior emotional experience rather than by present evidence alone. Third, it creates a mechanism for affective continuity, allowing the system to interpret new observations in light of historically accumulated emotional development rather than as isolated local events.

Retrieval is particularly important when current evidence is incomplete, ambiguous, or potentially misleading. A weak textual cue may become more interpretable when retrieved memory suggests a continuing pattern of frustration; a seemingly neutral visual signal may take on different significance if the system retrieves prior traces of hesitation or emotional suppression; an apparently salient acoustic cue may be down-weighted if retrieval indicates that it conflicts with a more stable longer-term affective trajectory. In this way, retrieval does not simply expand context length; it supports memory-guided reinterpretation of the present.

The output of this retrieval mechanism is not yet the final affective decision. Rather, it provides the historically grounded memory context that will be combined with current multimodal evidence in the next stage of the framework. The role of retrieval is therefore preparatory but decisive: it makes long-term emotional memory operational for present inference and provides the basis for the dynamic fusion strategies described in the following section.

\subsection{Dynamic Fusion Strategies}

Once historically relevant affective memory has been retrieved from the Memory Bear AI memory layer, the next step is to integrate this memory context with current multimodal evidence in a principled manner. In the Memory Bear AI framework, this process is not treated as a simple concatenation of present features and retrieved memory. Instead, the system adopts a \emph{dynamic fusion strategy} in which multimodal interpretation is calibrated by both current signal reliability and historically relevant affective context \cite{xue2025mmrc}.

This design follows the second core principle introduced earlier in the report: present multimodal interpretation should be memory-calibrated. In conventional multimodal pipelines, fusion is often driven by local feature interaction, learned attention weights, or the relative strength of currently available signals. While such mechanisms may work well under controlled conditions, they remain vulnerable when present cues are ambiguous, conflicting, noisy, or partially missing. Under these circumstances, a purely input-driven fusion process may over-trust a salient but unreliable modality, or underuse a weak but historically meaningful cue. The Memory Bear AI engine addresses this limitation by allowing retrieval-activated affective memory to participate directly in the calibration of multimodal fusion \cite{li2024multimodalalignmentfusion}.

At the architectural level, the dynamic fusion stage integrates two sources of evidence. The first is the current multimodal observation, including textual, acoustic, and visual affective representations. The second is the retrieved historical affective memory returned by the memory-driven retrieval mechanism. The role of fusion is to determine how these sources should be combined so that present interpretation remains sensitive to current evidence while also grounded in historically relevant emotional context.

A key idea in this stage is that modality contribution should not be determined solely by instantaneous feature salience. Instead, each modality should be evaluated along at least two dimensions: \emph{current reliability} and \emph{memory consistency}. Current reliability refers to the quality and interpretability of the present modality signal. For example, speech may be noisy, visual cues may be partially occluded, or language may be indirect and emotionally restrained. Memory consistency refers to the degree to which the present modality aligns with historically relevant affective traces retrieved from long-term memory. A modality that is only moderately strong in the current moment may become highly informative if it matches a stable emotional trajectory stored in memory, whereas a seemingly salient modality may deserve lower weight if it conflicts with previously accumulated affective evidence.

Let the modality-specific current representations be denoted abstractly as textual, acoustic, and visual affective evidence, and let \(M_t\) denote the retrieved historical affective memory. The dynamic fusion process can be expressed at a high level as:
\[
F_t = \mathrm{Fuse}(X_t, M_t)
\]
where \(X_t\) denotes the set of current multimodal inputs and \(F_t\) denotes the memory-calibrated multimodal representation used for downstream decision-making. More specifically, the contribution of each modality may be modulated by a weighting function:
\[
\alpha_i = f(r_i, s_i)
\]
where \(\alpha_i\) denotes the contribution weight of modality \(i\), \(r_i\) denotes its current reliability, and \(s_i\) denotes its consistency with retrieved affective memory. At the architectural level, the precise implementation of \(f(\cdot)\) may vary; the key point is that the proposed engine does not treat multimodal fusion as a function of present input alone.

This design yields several practical benefits. First, it improves interpretive stability in the presence of noisy or degraded modalities. If one modality is unreliable in the current interaction, the system can reduce its influence rather than allowing it to dominate the fused representation. Second, it allows weak but contextually meaningful signals to be amplified when they align with historically relevant emotional memory. Third, it supports disambiguation under modality conflict by using memory as a calibration reference rather than forcing all interpretation to be resolved within the local perceptual window.

The importance of memory-guided fusion becomes especially clear in affectively ambiguous settings. A weak textual cue may become more informative when it is consistent with a previously retrieved pattern of frustration or hesitation. A neutral facial expression may not be interpreted as truly neutral if the system retrieves a longer-term history of emotional suppression. Conversely, an apparently strong acoustic signal may be down-weighted if it conflicts with a more stable historically grounded emotional trajectory. In such cases, fusion is not simply an operation over concurrent signals; it becomes a context-sensitive process of affective reinterpretation.

The output of this stage is a memory-calibrated affective representation that is then passed to the final decision and updating stage. This fused representation is then passed to the final stage of the framework, where affective classification, decision-making, and post-inference memory updating are performed.

\subsection{Classification, Decision-Making, and Memory Updating}

The final stage of the Memory Bear AI engine transforms the memory-calibrated multimodal representation into an affective decision and then uses this decision to update the state of the memory system. This stage completes the transition from perception and memory-guided interpretation to decision-oriented affective understanding. Importantly, the decision process is not treated as a terminal endpoint. Instead, the inferred affective result is fed back into the broader memory lifecycle, allowing the system to strengthen, revise, reprioritize, or selectively forget emotional traces in light of newly interpreted evidence.

This stage serves two closely related purposes. First, it produces the task-level affective output required for downstream use, such as emotion classification, dimensional affect estimation, or a decision-oriented affective state for interaction support. Second, it closes the memory loop by allowing present interpretation to influence future reasoning. In this way, classification and decision-making are embedded in an evolving memory system rather than standing outside it.

\subsubsection{Affective Classification and Decision Layer}

The immediate function of the decision layer is to map the memory-enhanced multimodal representation produced by the dynamic fusion stage to a final affective output. Depending on the task setting, this output may take the form of a discrete emotion category, a continuous affective estimate, or a higher-level interaction state used to guide subsequent system behavior. The key architectural point is that this decision is not derived from current multimodal evidence alone, but from a representation that has already been calibrated by historically relevant affective memory.

Let the output of the dynamic fusion stage be denoted by \(F_t\). The decision layer can then be written at a high level as:
\[
y_t = \mathrm{Decide}(F_t)
\]
where \(y_t\) denotes the final affective output at time \(t\). In a categorical setting, \(y_t\) may correspond to an emotion label such as anger, sadness, neutrality, or joy. In a dimensional setting, it may correspond to a continuous affective estimate such as valence and arousal. More generally, \(y_t\) can also be interpreted as a decision-oriented affective state that supports downstream interaction, including response adjustment, escalation detection, or emotional state tracking across sessions.

The decision layer operates on a representation that has already been shaped by retrieved affective memory. This has two important implications. First, local ambiguity may be reduced before the final decision is made, because retrieved emotional history and memory-calibrated fusion have already contributed to interpretation. Second, the final output carries greater longitudinal meaning than a purely local label, since it is produced in the context of prior affective development rather than from an isolated interaction slice.

In practical terms, this means that the same present input may lead to different final decisions depending on the relevant emotional memory activated by the system. A brief neutral utterance following repeated signals of disappointment may be interpreted as resignation, whereas the same utterance following successful resolution of earlier tension may be interpreted as relief or acceptance. The decision layer therefore does not simply assign a label to a fused feature vector; it produces an affective judgment grounded in both present evidence and historically relevant emotional context.

At the learning level, this stage may be optimized with standard task objectives such as classification loss or regression loss, depending on the affective formulation adopted in the experimental setting. However, from the perspective of the overall Memory Bear AI engine, the decision layer should be understood less as an isolated prediction head and more as the point at which memory-guided interpretation is converted into an actionable affective state. This distinction is important because the inferred result is subsequently used to influence memory management in the next component of the framework.

The final affective output is subsequently incorporated into memory lifecycle management. The next subsection describes how newly inferred affective evidence is used for selective updating, forgetting, merging, and conflict resolution.

\subsubsection{Memory Lifecycle Management: Forgetting, Updating, and Conflict Resolution}

A memory-centered affective system requires explicit lifecycle management rather than passive accumulation. If every affective trace were retained indefinitely and with equal importance, the memory store would gradually become saturated with redundant, low-value, or outdated emotional evidence, reducing both interpretability and retrieval quality. For this reason, the final component of the Memory Bear AI engine is a memory lifecycle management mechanism that governs how emotional memory is updated after inference. This mechanism includes selective forgetting, priority adjustment, memory merging, memory revision, and conflict resolution.

The need for such a mechanism follows directly from the design principles introduced earlier in this report. Emotional memory should be selectively consolidated rather than indiscriminately accumulated, and historically relevant traces should remain retrievable without allowing obsolete or weakly informative memory to dominate later reasoning. Accordingly, the role of memory lifecycle management is not simply to delete old entries, but to maintain a useful, adaptive, and interpretable affective memory system over time.

Let the current long-term memory store be denoted by \(\mathcal{M}^{\mathrm{LTM}}_t\), and let \(y_t\) denote the newly inferred affective decision at time \(t\). The post-inference memory management process can be written abstractly as:
\[
\mathcal{M}^{\mathrm{LTM}}_{t+1} = \mathrm{Update}\!\left(\mathcal{M}^{\mathrm{LTM}}_t, y_t\right)
\]
where the update operator is understood broadly to include insertion, reinforcement, decay, merging, revision, and selective forgetting. At the architectural level, the precise implementation may vary, but the essential point is that affective memory is continuously reorganized in light of newly interpreted evidence rather than treated as a static archive.

\begin{figure}[H]
    \centering
    \includegraphics[width=0.8\linewidth]{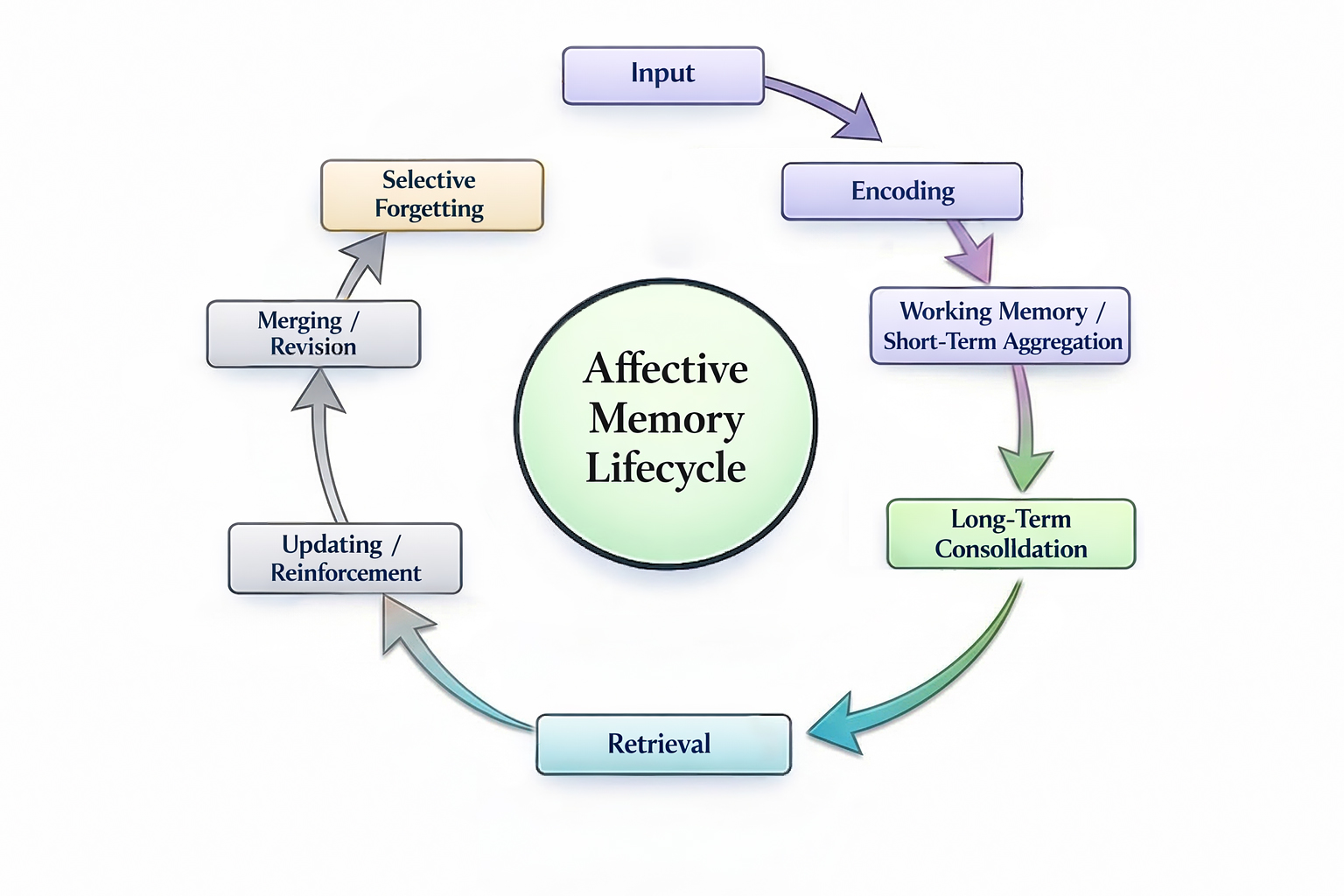}
    \caption{Conceptual illustration of the affective memory lifecycle in the Memory Bear AI framework. Affective information is continuously encoded, aggregated, consolidated, retrieved, updated, revised, and selectively forgotten rather than stored as a static archive.}
    \label{fig:memory_lifecycle}
\end{figure}

A first function of this mechanism is \emph{time-sensitive forgetting}. Emotional traces that remain unactivated for extended periods, that carry weak affective salience, or that prove to have little relevance for later interpretation should gradually lose priority. This decay process prevents the long-term memory from being dominated by stale local observations and helps preserve retrieval efficiency. In the present framework, forgetting is not treated as a failure of memory, but as a necessary condition for adaptive memory management.

A second function is \emph{selective strengthening and reprioritization}. Not all memories decay at the same rate. Emotionally intense, highly activated, or decision-relevant traces should retain higher importance and greater resistance to forgetting than weak, incidental, or low-value observations. In this sense, newly inferred affective decisions may reinforce previously stored emotional trajectories when they confirm or reactivate them. This allows the memory store to reflect not only what has been observed, but also which affective patterns have proved persistently relevant.

A third function is \emph{memory merging}. In extended interaction, multiple emotional traces may reflect the same evolving affective tendency rather than distinct independent events. If stored separately without control, such traces may create unnecessary redundancy and reduce the coherence of later retrieval. The Memory Bear AI framework therefore allows highly similar or repeatedly co-activated emotional memories to be merged into more compact and stable representations. This process preserves continuity while reducing unnecessary duplication.

A fourth function is \emph{memory revision}. Emotional interpretation is not always final at the moment it is first formed. New evidence may reveal that an earlier affective state was misunderstood, incomplete, or contextually misleading. For example, what initially appeared to be mild neutrality may later be better understood as emotional suppression; what appeared to be frustration may later be reinterpreted as anxiety or hesitation. In such cases, memory management should support revision rather than merely appending a new and contradictory trace on top of the old one.

A fifth function is \emph{conflict resolution}. In long-horizon interaction, newly inferred affective evidence may conflict with previously consolidated emotional memory. Such conflict should not be handled through simple accumulation alone. Instead, the system should assess whether the new trace reflects temporary deviation, correction of an earlier interpretation, or genuine emotional transition. This requires the memory layer to consider factors such as recency, salience, repeated activation, and contextual consistency when deciding whether to preserve both traces, reprioritize one over the other, or partially revise the prior memory state.

These operations make memory evolution an integral part of affective intelligence. The memory store is not merely extended after each interaction; it is reorganized. Some traces are strengthened, some weakened, some merged, some revised, and some forgotten. This ongoing reorganization ensures that the memory layer remains both historically grounded and dynamically adaptive.

In this way, the final stage of the Memory Bear AI engine closes the loop between present interpretation and future reasoning. Affective decisions do not terminate at output generation; they reshape the memory substrate that will later calibrate new multimodal evidence. Memory lifecycle management therefore provides the final component needed to transform the engine from a local multimodal classifier into a persistent affective system with continuity, prioritization, and adaptive evolution.

\section{Experimental Validation and Case-Based Analysis}

\subsection{Experimental Setup}

The experimental design is intended to evaluate whether a memory-centered multimodal architecture improves affective judgment in settings where historical context, modality reliability, and interaction continuity materially influence interpretation. The purpose of this report is therefore not to present Memory Bear AI as a generic benchmark-optimized emotion classifier, but to examine whether structured affective memory provides measurable value in practical multimodal inference.

Accordingly, the evaluation combines public multimodal datasets with a business-grounded internal dataset. The public datasets are used as reference settings for general multimodal affective modeling, while the main empirical emphasis is placed on a real-world Memory Bear AI dataset collected from practical application scenarios. This design allows the evaluation to cover both standard benchmark comparison and deployment-relevant conditions in which emotional meaning may be weakly expressed, temporally distributed, or partially obscured by missing or degraded modalities.

The following subsections describe the datasets, comparison settings, and evaluation metrics used in this study.

\subsubsection{Datasets}

The evaluation uses both public datasets and a business-grounded internal dataset. Public datasets are included primarily as reference settings for general multimodal affective modeling, whereas the main evaluation emphasis is placed on a real-world dataset collected from Memory Bear AI application scenarios.

\paragraph{IEMOCAP.}
We use IEMOCAP as a controlled public benchmark for multimodal emotion recognition \cite{busso2008iemocap}. Its synchronized text, speech, and visual signals provide a relatively clean evaluation setting, making it useful as a reference point for basic affective modeling performance.

\paragraph{CMU-MOSEI.}
We further include CMU-MOSEI as a more realistic public multimodal dataset \cite{zadeh2018cmumosei}. Compared with IEMOCAP, CMU-MOSEI contains more heterogeneous and naturally occurring multimodal expressions, and is therefore used as a second reference setting for evaluating generalization beyond tightly controlled benchmark conditions.

\paragraph{Memory Bear AI real-world evaluation dataset.}
The main evaluation focus of this report is a real-world internal dataset collected from Memory Bear AI business scenarios. This dataset contains multimodal interaction records from practical application settings and covers multiple usage contexts with heterogeneous interaction patterns. More importantly, it naturally includes conditions that are central to the motivation of this work, such as incomplete modalities, uneven signal quality, partial observations, and scenario-dependent variation in emotional expression.

This dataset is particularly important because the proposed framework is intended to evaluate the value of memory in affective judgment rather than to optimize only for benchmark-style classification. In practical interactions, emotional meaning is often distributed across time, weakly expressed in the current turn, or partially obscured by modality limitations\cite{pham2026missbench,wu2024missingmodalitysurvey}. For this reason, the Memory Bear AI dataset serves as the primary evaluation resource for testing whether structured emotional memory, retrieval, and memory-guided fusion improve robustness and continuity in realistic affective inference.

Overall, the public datasets act as reference settings, whereas the Memory Bear AI real-world dataset functions as the main test bed for practical validation.

\subsubsection{Comparison Settings}

To evaluate the contribution of the proposed memory-centered design, we compare the Memory Bear AI framework against several categories of multimodal affective systems rather than against a single model family. This comparison strategy reflects the fact that the proposed framework is not merely a fusion module, but a broader architectural design that combines multimodal representation learning, structured affective memory, memory-guided retrieval, and decision-dependent memory updating.

The comparison settings are organized into four groups.

\paragraph{(1) Conventional multimodal fusion baselines.}
The first group consists of conventional multimodal baselines that combine textual, acoustic, and visual information through standard fusion strategies. These baselines represent the classical formulation of multimodal emotion recognition, in which modality-specific features are integrated locally without explicit long-horizon memory support.

\paragraph{(2) Strong neural multimodal baselines.}
The second group consists of stronger neural architectures that model cross-modal interaction through attention mechanisms, hierarchical fusion, or transformer-style designs. These systems are more expressive than conventional fusion baselines and better reflect the current state of multimodal affective modeling.

\paragraph{(3) Context- or memory-aware comparison models.}
The third group consists of comparison systems that incorporate contextual history or memory-like modeling in some form, such as recurrent context tracking, memory-network-style processing, or conversational state modeling. This group is particularly important because the proposed framework is explicitly memory-centered. Comparing against these models helps clarify whether the gains of Memory Bear AI arise from memory in general or from its more structured design, including affective memory formation, short-term aggregation, long-term consolidation, retrieval, and lifecycle management.

\paragraph{(4) Internal ablated variants of the proposed framework.}
The fourth group consists of internal architectural variants obtained by removing or simplifying specific components of the Memory Bear AI engine. These variants are used to isolate the contribution of the memory-centered design itself. In particular, we consider variants without structured emotional memory formation, without memory-driven retrieval, without memory-guided fusion, without post-inference memory updating, and without the long-term memory branch.

Overall, the comparison setting serves two purposes: first, to compare the proposed framework with representative multimodal baselines at the performance level; and second, to examine the contribution of its memory-centered components through internal ablations. This design allows the evaluation to reflect both empirical effectiveness and the specific value of the Memory Bear AI architecture.

\subsubsection{Evaluation Metrics}

The evaluation focuses on three aspects: overall predictive performance, class-balanced performance, and robustness under imperfect multimodal conditions.

\paragraph{Predictive performance.}
For categorical affective tasks, we report Accuracy as the primary measure of overall predictive correctness.

\paragraph{Class-balanced performance.}
Because affective datasets are often imbalanced across categories, we additionally report Weighted F1 and Macro F1 whenever applicable. These metrics provide a more balanced view of classification quality across frequent and infrequent emotion classes.

\paragraph{Robustness under imperfect conditions.}
For the real-world Memory Bear AI dataset in particular, we also evaluate performance under heterogeneous modality conditions, including missing modalities and unstable signal quality. In these settings, the goal is not only to measure absolute task performance, but also to assess performance retention when the input departs from the ideal complete-modality condition.

These metrics allow us to evaluate both prediction quality and robustness under practical multimodal conditions.

\subsection{Main Experimental Findings}

Across benchmark, public, and business-grounded evaluation settings, the experimental results reveal a consistent pattern: the proposed framework is especially advantageous in scenarios where affective interpretation depends not only on the current local multimodal segment, but also on continuity, historical reinterpretation, and robustness under imperfect evidence conditions.

Table~\ref{tab:long_horizon_results} presents the overall comparison across three datasets. On the two public datasets, the proposed framework performs competitively and remains above the strongest comparison models. The improvement is moderate on IEMOCAP and CMU-MOSEI, which is expected because these datasets mainly function as reference benchmarks rather than as dedicated long-horizon emotional memory evaluation settings. In contrast, the relative advantage becomes clearer on the Memory Bear AI Business Dataset. Although the absolute scores on this dataset are lower for all systems, the proposed framework still remains consistently above the comparison baselines, suggesting that structured emotional memory is particularly useful when multimodal evidence is weaker, noisier, and less self-sufficient.

This pattern is consistent with the intended role of the architecture. The value of the proposed system does not come only from stronger local multimodal fusion, but from its ability to preserve, reactivate, and reorganize historically relevant affective evidence. As a result, the framework shows its clearest benefit when emotional meaning cannot be determined reliably from the current segment alone and when interpretation must be supported by accumulated affective context.

A second noteworthy pattern concerns class-balanced performance. In addition to improvements in Accuracy and Weighted F1, the proposed framework also maintains the strongest Macro F1 across all three datasets. This suggests that the benefit of Memory Bear AI is not limited to dominant or frequent affective categories, but extends to a more balanced category-level performance profile, including cases that are relatively infrequent or harder to classify. Importantly, this should be interpreted as stronger balance across emotion categories rather than as evidence that every individual modality performs better in isolation.

Overall, the cross-dataset comparison indicates that the proposed framework is effective in both standard benchmark settings and more deployment-relevant business scenarios, while its relative advantage becomes more visible in the latter.

\begin{table*}[t]
\centering
\caption{Overall comparison across three datasets.}
\label{tab:long_horizon_results}
\resizebox{\textwidth}{!}{
\begin{tabular}{llccc}
\hline
Model & Dataset & Accuracy & Weighted F1 & Macro F1 \\
\hline

Traditional fusion baseline (LF-LSTM)
& \multirow{4}{*}{IEMOCAP}
& 71.8
& 49.5
& 46.4 \\

Strong neural multimodal baseline (MulT)
&
& 77.6
& 56.9
& 53.7 \\

Context-/emotion-aware baseline (EmoEmbs)
&
& 72.0
& 49.8
& 46.7 \\

Memory Bear AI
&
& \textbf{78.8}
& \textbf{57.3}
& \textbf{54.2} \\
\hline

Traditional fusion baseline (LF-LSTM)
& \multirow{4}{*}{CMU-MOSEI}
& 63.1
& 43.3
& 39.5 \\

Strong neural multimodal baseline (MulT)
&
& 65.4
& 45.2
& 41.8 \\

Context-/emotion-aware baseline (EmoEmbs)
&
& 64.2
& 44.2
& 40.8 \\

Memory Bear AI
&
& \textbf{66.7}
& \textbf{45.8}
& \textbf{42.3} \\
\hline

Traditional fusion baseline
& \multirow{4}{*}{Memory Bear AI Business Dataset}
& 60.2
& 41.7
& 37.8 \\

Strong neural multimodal baseline
&
& 63.8
& 44.5
& 40.9 \\

Context-/memory-aware baseline
&
& 65.1
& 46.0
& 42.7 \\

Memory Bear AI
&
& \textbf{68.4}
& \textbf{48.6}
& \textbf{45.9} \\
\hline
\end{tabular}
}
\end{table*}

\subsection{Mechanism and Ablation Analysis}

To better understand why the proposed framework outperforms comparison systems, we next examine both mechanism-level evidence and internal ablation results. Table~\ref{tab:dynamic_fusion_results} focuses on progressively enriched model variants, whereas Table~\ref{tab:ablation_results} isolates the contribution of core architectural components.

Table~\ref{tab:dynamic_fusion_results} shows that the gain of Memory Bear AI is not simply a consequence of adding more model capacity, but emerges through a structured progression from unimodal inference to multimodal fusion, temporal-context modeling, memory-guided fusion, and finally the full system. Two patterns are especially important. First, moving from the best single-modality baseline to vanilla multimodal fusion yields a clear gain on both datasets, confirming the value of multimodal evidence integration itself. Second, the gains continue as memory mechanisms are introduced: adding temporal context improves performance further, and introducing memory-guided fusion yields another consistent increase. The full Memory Bear AI system remains the strongest setting on both datasets, indicating that memory-guided fusion is most effective when it operates together with structured memory formation and memory lifecycle management rather than as an isolated retrieval add-on.

The gain is particularly visible on the business-grounded dataset. On this dataset, the difference between vanilla multimodal fusion and memory-guided fusion is noticeably larger than on CMU-MOSEI, and the full system continues to improve on top of both. This suggests that the practical value of memory-guided fusion becomes clearer when modality quality is uneven, when current cues are ambiguous, and when interaction history contains meaningful affective regularities that cannot be recovered from local evidence alone. The corresponding gains in Macro F1 further indicate that this benefit is associated not only with higher aggregate performance, but also with a more balanced category-level interpretation under more realistic conditions.

\begin{table*}[t]
\centering
\caption{Effect of memory-guided fusion on multimodal affective interpretation across representative datasets.}
\label{tab:dynamic_fusion_results}
\resizebox{\textwidth}{!}{
\begin{tabular}{llccc}
\hline
Model Variant & Dataset & Accuracy & Weighted F1 & Macro F1 \\
\hline

Best single-modality baseline
& \multirow{5}{*}{CMU-MOSEI}
& 61.7
& 41.0
& 39.6 \\

Vanilla multimodal fusion
&
& 64.0
& 43.8
& 41.2 \\

Multimodal + temporal context
&
& 65.2
& 44.7
& 41.8 \\

Multimodal + memory-guided fusion
&
& 66.0
& 45.3
& 42.1 \\

Memory Bear AI
&
& \textbf{66.7}
& \textbf{45.8}
& \textbf{42.3} \\
\hline

Best single-modality baseline
& \multirow{5}{*}{Memory Bear AI Business Dataset}
& 57.6
& 39.5
& 35.8 \\

Vanilla multimodal fusion
&
& 60.2
& 41.7
& 37.8 \\

Multimodal + temporal context
&
& 62.7
& 43.8
& 40.2 \\

Multimodal + memory-guided fusion
&
& 66.1
& 46.8
& 43.5 \\

Memory Bear AI
&
& \textbf{68.4}
& \textbf{48.6}
& \textbf{45.9} \\
\hline
\end{tabular}
}
\end{table*}

Table~\ref{tab:ablation_results} further clarifies which components are responsible for the observed gains. Several patterns are clear. Removing structured emotional memory formation leads to the largest overall drop, indicating that the gains of the proposed framework depend not only on retaining context, but on organizing multimodal affective evidence into explicit memory structures. Removing memory-driven retrieval also produces a substantial decline, showing that longer-horizon affective memory is useful only when it can be selectively reactivated during current inference. The variant without memory-guided fusion also performs clearly worse than the full model, confirming that an important part of the gain comes from calibrating present interpretation against retrieved emotional memory rather than relying on local fusion alone.

The effect of removing post-inference memory updating is smaller but still meaningful. This suggests that lifecycle management contributes less to immediate classification performance than memory formation or retrieval, but remains important for longer-horizon continuity and adaptive memory evolution. A similar pattern is observed when the long-term memory branch is removed: the model retains some benefit from shorter-range contextual aggregation, but loses the broader continuity that supports more stable affective interpretation across interaction segments.

Taken together, the mechanism comparison and ablation results support the claim that the proposed framework should be understood as a coordinated architecture rather than as a collection of loosely connected modules. The strongest performance is obtained only when structured memory formation, retrieval, memory-guided fusion, updating, and longer-range memory support are all present.

\begin{table*}[t]
\centering
\caption{Ablation study of the Memory Bear AI framework on the Memory Bear AI Business Dataset.}
\label{tab:ablation_results}
\resizebox{\textwidth}{!}{
\begin{tabular}{lccc}
\hline
Model Variant & Accuracy & Weighted F1 & Macro F1 \\
\hline

Memory Bear AI (full)
& \textbf{68.4}
& \textbf{48.6}
& \textbf{45.9} \\

w/o memory formation
& 63.1
& 43.9
& 40.6 \\

w/o memory retrieval
& 64.0
& 44.8
& 41.5 \\

w/o memory-guided fusion
& 64.8
& 45.4
& 42.1 \\

w/o memory updating
& 67.1
& 47.4
& 44.6 \\

w/o long-term memory branch
& 64.3
& 45.0
& 41.8 \\
\hline
\end{tabular}
}
\end{table*}

\subsection{Robustness Under Noise and Missing Modalities}

We further examine the robustness of the proposed framework under heterogeneous multimodal conditions, with particular attention to missing modalities, unstable signal quality, and business-grounded deployment scenarios. The purpose of this analysis is to determine whether the memory-centered design improves performance retention when the input departs from the ideal complete-modality setting.

\begin{figure}[t]
    \centering
    \includegraphics[width=\linewidth]{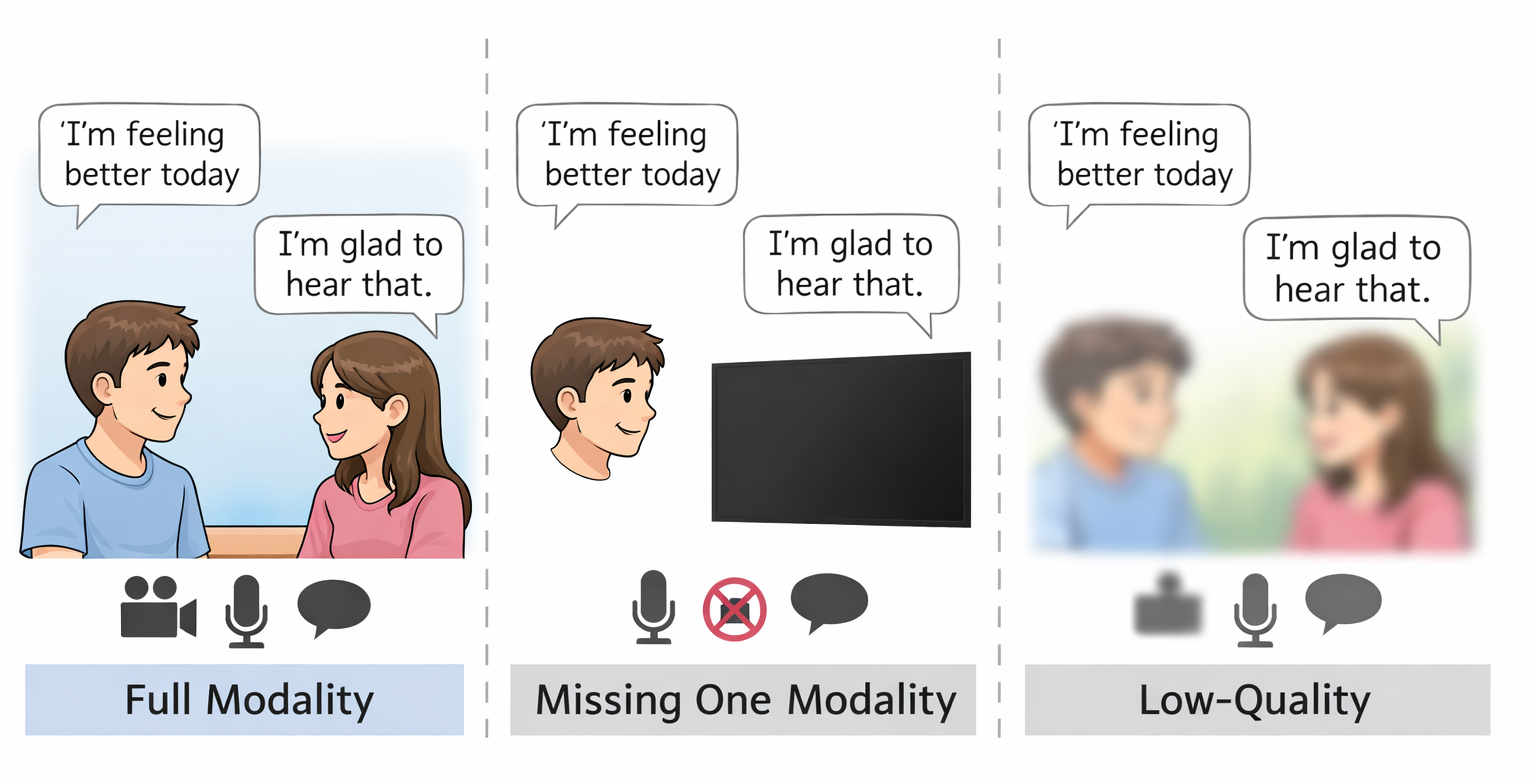}
    \caption{Illustration of the three modality conditions used in robustness evaluation: \textbf{Complete}, where textual, acoustic, and visual signals are all available; \textbf{Missing One}, where one channel is absent; and \textbf{Low Quality}, where multimodal input is available but degraded in reliability.}
    \label{fig:modality_conditions}
\end{figure}

Figure~\ref{fig:modality_conditions} illustrates the three modality conditions used in this robustness analysis: complete-modality input, one-modality-missing input, and low-quality multimodal input. As shown in Table~\ref{tab:robustness_results}, the Memory Bear AI engine maintains stronger performance than all comparison systems under both missing-modality and low-quality-signal conditions. This advantage is especially visible when the remaining available evidence is insufficient for reliable local interpretation on its own. In such cases, the proposed framework benefits from its ability to reactivate historically relevant affective memory and to use that memory to calibrate current multimodal interpretation.

\begin{table*}[t]
\centering
\caption{Robustness comparison on the Memory Bear AI Business Dataset under heterogeneous modality conditions. All values denote overall classification accuracy. Performance retention is calculated relative to the complete-modality setting.}
\label{tab:robustness_results}
\resizebox{\textwidth}{!}{
\begin{tabular}{lcccc}
\hline
Model & Complete & Missing One & Low Quality & Retention (\%) \\
\hline

Traditional Fusion Baseline
& 60.2
& 54.1
& 51.6
& 87.8 \\

Strong Neural Multimodal Baseline
& 63.8
& 58.7
& 56.2
& 90.2 \\

Context-/Memory-Aware Baseline
& 65.1
& 60.5
& 58.1
& 90.9 \\

Memory Bear AI
& \textbf{68.4}
& \textbf{64.2}
& \textbf{62.0}
& \textbf{92.3} \\
\hline
\end{tabular}
}
\end{table*}

Several patterns are noteworthy. First, the robustness gain is visible not only when a modality is absent, but also when multimodal input remains available while its quality is degraded. This suggests that the advantage of the framework does not come only from multimodal redundancy, but from the interaction between retrieval and memory-guided fusion under uncertain evidence conditions. Second, the proposed system shows the strongest performance retention across heterogeneous business scenarios, indicating that the memory layer remains useful even when modality quality varies naturally rather than through artificially simplified degradation alone. Third, the performance gap between the proposed framework and comparison models widens as input conditions become less reliable, which is consistent with the intended role of memory-guided affective calibration.

The retention results are especially informative in this respect. While all systems degrade under missing or low-quality modalities, the proposed framework preserves the largest proportion of its complete-condition performance. This indicates that structured emotional memory contributes not only to stronger nominal accuracy, but also to greater resilience when multimodal evidence becomes incomplete, ambiguous, or unstable. In deployment-oriented settings, this property is particularly important because real-world interaction data are rarely complete and often contain uneven signal reliability across modalities.

Although the present robustness table reports overall accuracy, the same interpretation would be further strengthened if future experiments also report Weighted F1 or Macro F1 under degraded conditions. In particular, a smaller decline in Macro F1 would suggest that the proposed framework preserves not only overall predictive performance, but also more balanced category-level interpretation when multimodal evidence becomes unreliable.

These findings show that the framework's robustness is not limited to standard benchmark conditions. Its value becomes clearer when multimodal evidence is incomplete, uneven, or difficult to interpret locally. The business-grounded evaluation is especially informative in this respect because it contains naturally occurring variation in modality availability, signal quality, and interaction context. The fact that the proposed framework maintains stronger robustness under these conditions supports the practical relevance of the Memory Bear AI design.

Overall, the robustness results reinforce the broader conclusion that memory-centered affective modeling contributes not only to predictive performance, but also to stability under imperfect multimodal conditions. This property is essential for real-world multimodal affective systems and motivates the qualitative case analysis presented next.

\subsection{Representative Case Studies}

\begin{figure}[t]
    \centering
    \includegraphics[width=\linewidth]{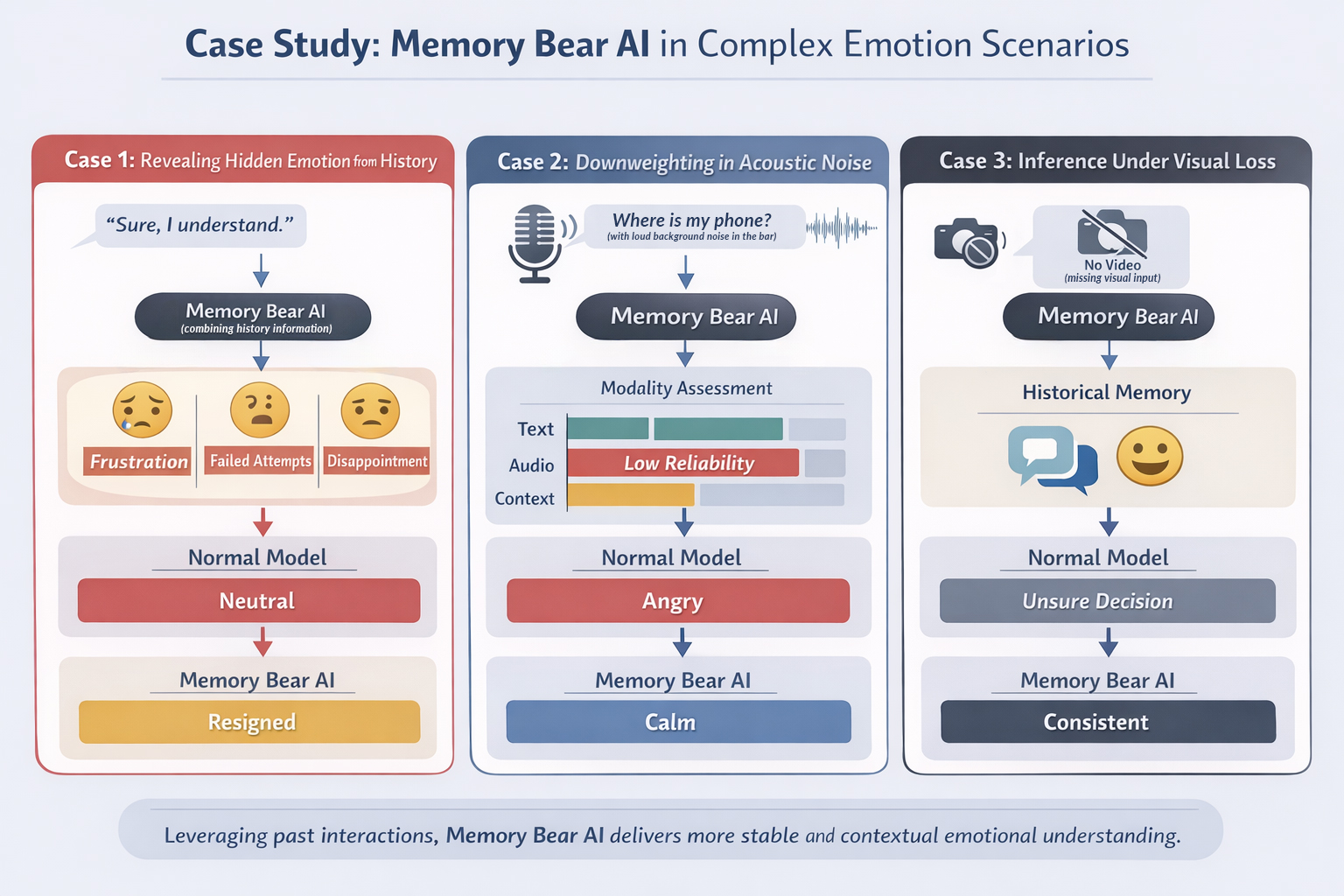}
    \caption{Illustration of three representative case studies for Memory Bear AI under complex affective inference scenarios. \textbf{Case 1} shows how historical affective trajectory helps reveal a hidden emotion that may appear neutral from the current utterance alone. \textbf{Case 2} shows how memory-guided fusion downweights unreliable acoustic evidence under strong background noise. \textbf{Case 3} shows how historical memory supports stable inference when visual input is partially or fully unavailable.}
    \label{fig:case_study_overview}
\end{figure}

To complement the quantitative results, we present three representative cases that illustrate how the Memory Bear AI framework behaves in challenging affective scenarios: \textbf{(1) hidden emotion revealed by historical affective trajectory}, \textbf{(2) automatic down-weighting under acoustic noise}, and \textbf{(3) stable inference under missing visual input}. As illustrated in Figure~\ref{fig:case_study_overview}, the representative cases consistently show that the advantage of Memory Bear AI becomes most visible when current multimodal evidence is weak, noisy, incomplete, or emotionally indirect.

\paragraph{Case 1: Historical affective trajectory reveals hidden emotion.}
In the first case, the user's current utterance is short, linguistically mild, and superficially non-emotional: \emph{``Okay, I understand.''} A conventional multimodal baseline tends to classify this input as neutral or slightly positive because the sentence contains no explicit negative wording, the facial signal is restrained, and the acoustic contour is relatively flat.

The Memory Bear AI framework produces a different interpretation. Through long-term consolidation and retrieval, the system has access to a recent interaction history marked by repeated disappointment, several unsuccessful attempts at resolution, and a gradual shift from direct complaint to more restrained language. When the current utterance is interpreted against this retrieved affective trajectory, it is no longer treated as neutral acknowledgment, but as resignation or suppressed frustration. This case therefore illustrates the core value of memory-centered affective inference: emotional meaning may be hidden at the current turn yet recoverable through historically grounded reinterpretation.

\paragraph{Case 2: Acoustic noise triggers automatic down-weighting.}
In the second case, the user is speaking in an environment with substantial background noise. The audio channel contains irregular spikes and unstable energy, and a conventional multimodal model overreacts to these acoustic artifacts, interpreting the utterance as anger or agitation. Because the acoustic modality appears salient in the current moment, a local fusion strategy assigns it excessive influence even though the underlying signal quality is poor.

The Memory Bear AI framework responds differently. The current acoustic channel is assessed as unreliable, while the textual content remains relatively calm and the retrieved affective memory indicates no prior escalation pattern in the surrounding interaction. During memory-guided fusion, the system therefore reduces the contribution of the acoustic modality instead of allowing it to dominate the final interpretation. The resulting judgment remains more stable and more consistent with the broader interaction context. This case highlights that the framework benefits not only from retrieval itself, but from adaptive calibration of multimodal evidence under imperfect sensing conditions.

\paragraph{Case 3: Missing visual input with memory-supported continuation.}
In the third case, the visual channel is partially unavailable because the camera feed is weak, occluded, or missing entirely. A comparison system that relies heavily on complete-modality local perception shows a marked drop in confidence, because the remaining textual and acoustic evidence is relatively subtle and does not strongly indicate a clear emotional category on its own.

The Memory Bear AI framework remains more stable. Before the visual modality disappears, the system has already encoded and consolidated affective information from the earlier interaction context. When the current turn is processed, retrieval reactivates this context, and the remaining textual and acoustic channels are interpreted against the retrieved memory rather than in isolation. As a result, the framework can preserve a coherent affective judgment even though one modality is no longer available. This case aligns with the robustness findings reported earlier and illustrates how memory continuity can compensate for incomplete real-time perception.

Across these cases, the same pattern appears repeatedly: the proposed framework is most useful when the current signal is weak, noisy, incomplete, or emotionally indirect. In such situations, structured memory, retrieval, and memory-guided fusion allow the system to produce a more coherent interpretation than local perception alone. The qualitative evidence therefore complements the quantitative results by showing how the proposed architecture behaves in scenarios that are practically important but difficult to capture through local multimodal classification alone.

\section{Capability Interpretation and Practical Value}

\subsection{Long-Horizon Emotional Dependency Modeling}

One of the main capabilities enabled by the Memory Bear AI framework is long-horizon emotional dependency modeling. In many real interactions, the meaning of the current emotional expression cannot be determined reliably from the current utterance, frame, or acoustic segment alone. It depends on how the interaction has evolved over time: whether frustration has accumulated, whether tension has eased, or whether a previously observed state is being sustained or reversed.

The proposed framework addresses this problem by treating memory as an explicit organizing layer for affective judgment. Instead of processing each interaction moment as an isolated prediction unit, the system retains and reuses historically relevant affective information through structured memory formation, short-term aggregation, long-term consolidation, and retrieval. This allows the current signal to be interpreted as part of an ongoing emotional trajectory rather than as a standalone event.

As a result, the system is better able to distinguish transient fluctuation from persistent tendency and to produce a more coherent view of how affect develops across interaction. This capability is one of the core advantages of the Memory Bear AI design and reflects the shift from local emotion recognition toward persistent affective understanding.

\subsection{Robustness Through Memory-Guided Calibration}

A second important capability of the Memory Bear AI framework is robustness under imperfect multimodal conditions. In practical settings, affective judgment often has to be made from signals that are noisy, uneven in quality, or partially unavailable. Under these conditions, systems that depend mainly on the current local input are more likely to produce unstable interpretation.

The proposed framework addresses this problem through memory-guided calibration. The system interprets the current multimodal observation against historically relevant affective memory, so that modality contribution depends not only on present signal strength, but also on reliability and consistency with prior emotional context.

This is useful in two common situations. The first is noisy input: when an acoustic or visual signal is unreliable, the framework can reduce its influence rather than allowing an anomalous cue to dominate the final interpretation. The second is modality incompleteness: when one channel is partially missing, the system can continue operating by combining the remaining evidence with retrieved affective memory from the ongoing interaction history.

The key point is that robustness here is not achieved only through multimodal redundancy, but through memory-aware interpretation. By combining structured memory, retrieval, and dynamic fusion, the Memory Bear AI engine maintains more stable affective judgment when the current evidence is noisy, incomplete, or difficult to interpret locally.

\begin{figure}[H]
    \centering
    \includegraphics[width=0.8\linewidth]{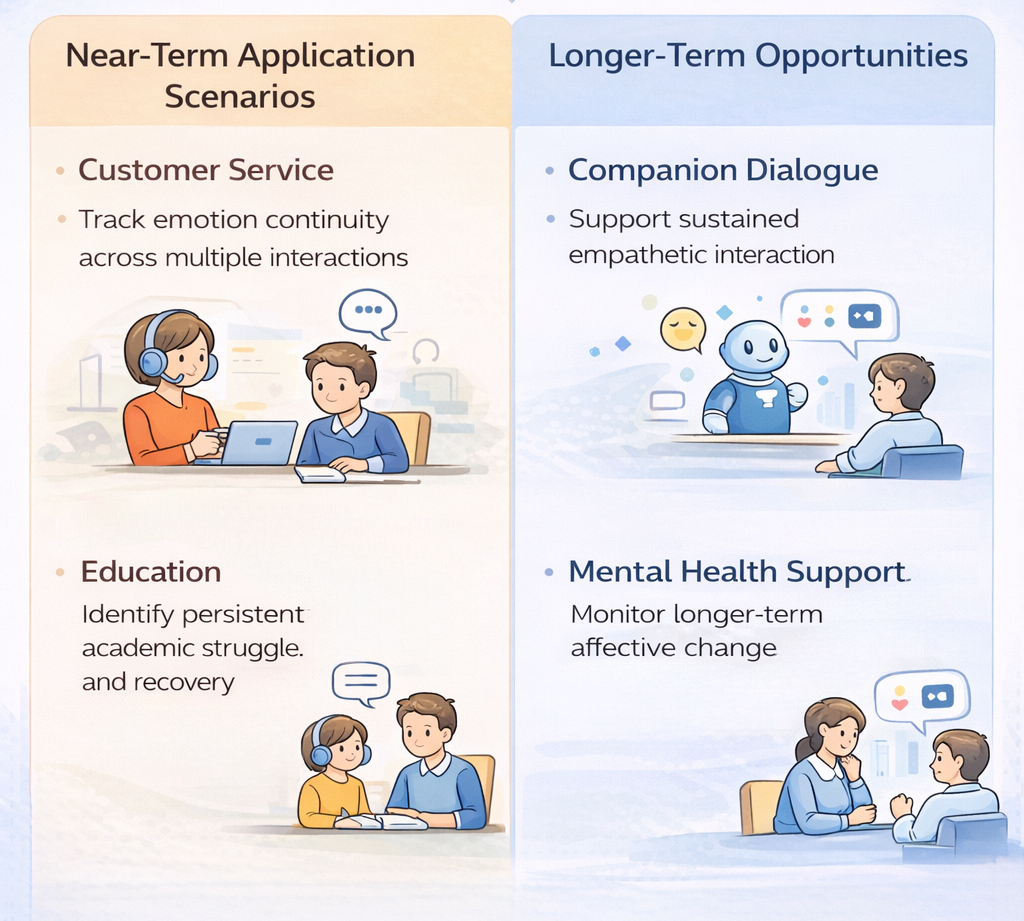}
    \caption{Capability Interpretation and Practical Value.}
    \label{fig:capability_practical}
\end{figure}

\subsection{Toward Persistent Personalization}

Beyond long-horizon dependency modeling and robustness, the Memory Bear AI framework also points toward a broader capability: persistent personalization in affective interaction. This is not the primary empirical claim of the present report, and it is not treated here as a fully validated result. Rather, it is a natural extension of the proposed architecture.

If affective information can be structured, retained, retrieved, and updated across interaction horizons, then the system can in principle accumulate a more stable understanding of user-specific emotional patterns over time. Such patterns may include recurring forms of frustration, typical responses to failure or success, and preferred styles of implicit or explicit emotional expression.

This does not imply complete or infallible personal understanding. Personalization in affective systems remains difficult, especially when emotional expression varies across context and situation. However, the architecture proposed in this report provides the structural conditions under which such personalization can become possible. For this reason, persistent personalization should be understood as an important future-facing capability of the Memory Bear AI framework rather than as a central validated result of the present study.

\subsection{Application Scenarios}

The capabilities described above are especially relevant in settings where affective judgment must remain stable across extended interaction, imperfect multimodal input, or repeated user engagement. In this report, the most immediate application scenarios are \textbf{customer service} and \textbf{education}, while several additional directions appear promising over a longer time horizon.

\paragraph{Near-term deployment priorities.}
In \emph{customer service}, user emotion is often expressed indirectly and unfolds across multiple turns. A memory-centered affective system can track emotional trajectory, reinterpret mild current responses in light of prior interaction, and remain more stable when modality quality is uneven.

In \emph{education}, affective judgment is also strongly trajectory-dependent. Confusion, hesitation, disengagement, and frustration often emerge gradually rather than as isolated events. A system that can retain and reuse affective context over time is therefore better suited to identifying persistent struggle, temporary uncertainty, or recovery from earlier difficulty.

\paragraph{Longer-term directions.}
Beyond these near-term priorities, the proposed framework also has potential value in \emph{companion dialogue}, \emph{enterprise intelligent assistants}, and \emph{mental health support settings}. In companion-style interaction, persistent affective memory may help maintain continuity across repeated conversations. In enterprise assistant settings, it may support longer-term user state awareness and more adaptive interaction. In mental health-related applications, the relevant value lies not in diagnosis, but in more stable longitudinal awareness of affective change under sustained interaction.

Overall, the practical value of the Memory Bear AI design lies in its ability to support continuity-aware affective judgment under realistic interaction conditions. The nearer-term emphasis of the present report is therefore on customer service and education, while the remaining scenarios are better understood as medium- to longer-term directions for further deployment and study.

\section{Limitations and Future Directions}

\subsection{Dataset and Evaluation Limitations}

Although the experimental results support the effectiveness of the proposed framework, several limitations of the current evaluation should be acknowledged.

First, existing public multimodal affective datasets are not specifically designed to evaluate long-horizon emotional memory\cite{lian2024merbench,maharana2024locomo,wu2024missingmodalitysurvey}. Datasets such as IEMOCAP and CMU-MOSEI are useful for benchmarking general multimodal affective modeling, but they do not fully reflect the longer interaction horizons, repeated user-specific trajectories, and persistent contextual dependencies that motivate the Memory Bear AI design. As a result, they can only partially capture the value of structured affective memory, especially in scenarios where emotional meaning depends on accumulation across time rather than on the current local segment alone.

Second, the current evaluation emphasizes practical robustness, but it still does not exhaust the full range of longitudinal affective behavior encountered in real deployment\cite{pham2026missbench,wu2024missingmodalitysurvey}. The internal Memory Bear AI dataset provides a more realistic test bed because it includes scenario variation, incomplete modalities, and uneven signal quality. However, it is still limited by the scope of available business scenarios, data coverage, and annotation conditions. It should therefore be viewed as a practical validation resource rather than as a definitive benchmark for persistent affective memory.

Third, some of the advantages claimed for the framework are easier to validate indirectly than directly. Improvements in long-horizon dependency modeling and robustness under degraded conditions can be observed through comparative performance and case analysis, but the deeper value of persistent affective memory would be better assessed with datasets explicitly constructed for longitudinal emotional continuity, retrieval relevance, and cross-session affective interpretation. Such evaluation resources remain limited at present.

For these reasons, the current results should be interpreted as strong evidence that memory-centered affective modeling is useful, but not as a complete evaluation of all the long-term capabilities implied by the architecture.

\begin{figure}[H]
    \centering
    \includegraphics[width=0.8\linewidth]{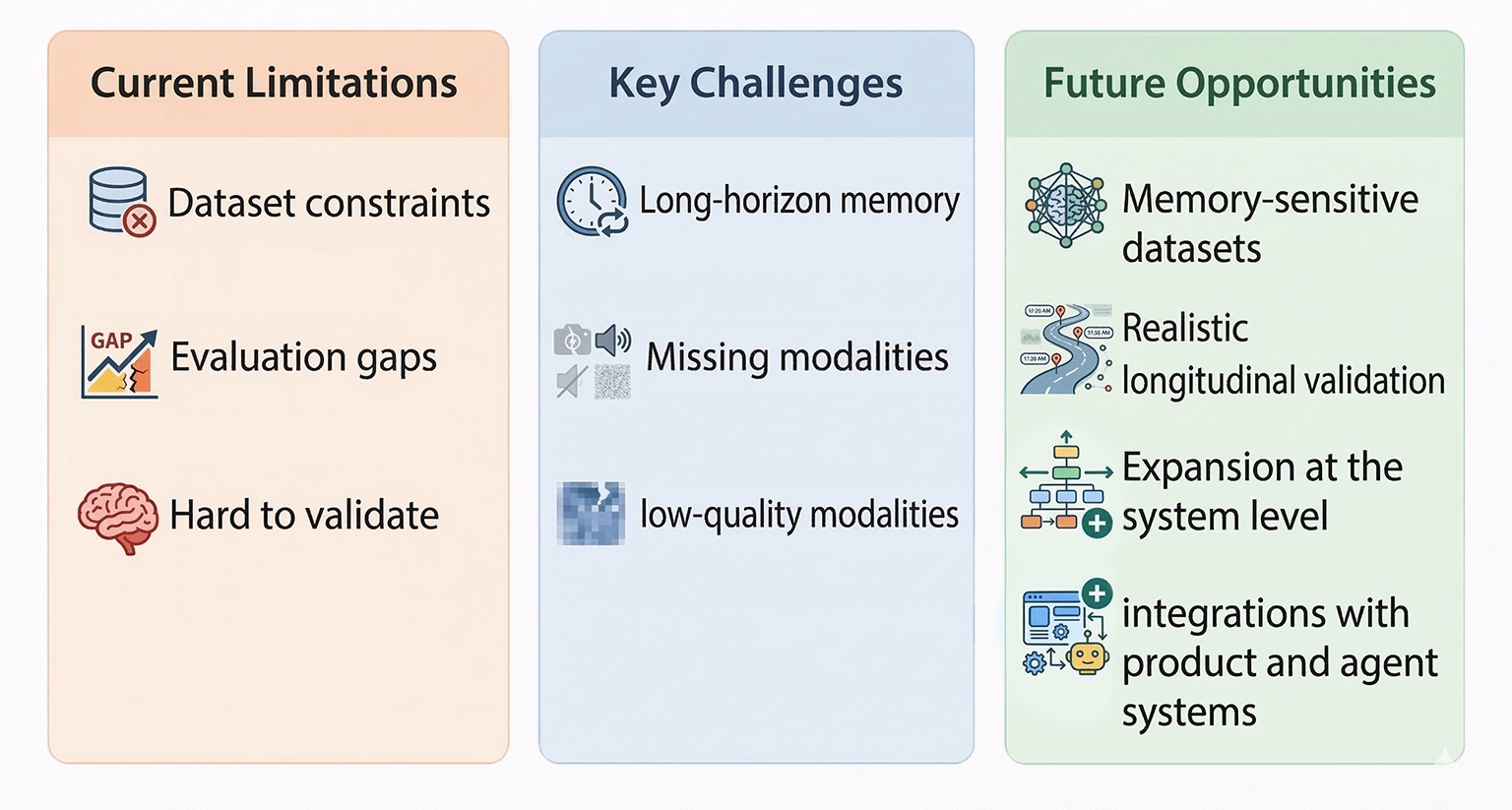}
    \caption{Limitations and Future Directions.}
    \label{fig:limit_directions}
\end{figure}

\subsection{Engineering Approximation of Memory Science}

A second limitation concerns the interpretation of \emph{memory science} in the present report. The Memory Bear AI framework is inspired by ideas from cognitive and affective memory research, but it should not be understood as a one-to-one simulation of human memory or brain-level emotional processing.

In this report, memory science is used primarily as an engineering design perspective. It provides a set of organizing principles for system construction, including structured encoding, selective retention, associative retrieval, contextual calibration, and adaptive updating. These principles are useful because they offer a more realistic way to model persistent affective judgment than purely local prediction. At the same time, they remain abstractions designed for computational systems rather than biological claims about how human memory operates in full detail.

This distinction matters for two reasons. First, it prevents overstatement. The proposed framework does not claim to reproduce the full mechanisms of human emotional memory, nor does it suggest that concepts such as working memory, consolidation, retrieval, or forgetting are implemented in exactly the same way as in human cognition. Second, it clarifies the contribution of the present work. The value of the Memory Bear AI design lies in showing that memory-inspired system structure can improve affective judgment in practice, not in asserting neuroscientific equivalence.

Accordingly, terms such as \emph{memory}, \emph{consolidation}, \emph{retrieval}, and \emph{forgetting} should be read in this report as functional engineering concepts. They describe how affective information is organized and managed within the proposed architecture, rather than claiming a literal replication of human memory processes.

\subsection{Future Development of the Memory Bear AI Memory Infrastructure}

The future development of the Memory Bear AI memory infrastructure should proceed along both evaluation and deployment directions. Among these, one of the most important next steps is the construction of datasets and benchmarks that can more directly reveal the value of emotional memory.

A first priority is therefore the development of \emph{memory-sensitive affective evaluation datasets}. Most existing public multimodal emotion datasets are useful for general multimodal affective modeling, but they do not fully capture the kinds of scenarios in which emotional memory becomes most important: gradual emotional drift, repeated affective patterns, context-dependent reinterpretation, cross-turn emotional continuity, and missing-modality conditions in which prior affective context must compensate for incomplete current input. A key future direction is therefore to create datasets that are explicitly structured to surface the contribution of emotional memory rather than only local emotion recognition. More broadly, this direction can also be understood as part of a larger transition from
memory-augmented interaction to memory-grounded cognition in AI systems, as argued
in the broader Memory Bear framework \cite{wen2025memorybear}.

Such datasets should include interaction sequences in which the current emotional meaning cannot be reliably inferred from the present local signal alone. Instead, accurate interpretation should depend on prior emotional trajectory, repeated contextual activation, accumulated user state, or retrieval of historically relevant affective traces. They should also include naturally occurring or deliberately designed cases involving noisy signals, modality degradation, and modality absence, so that the role of memory in stabilizing affective judgment can be evaluated more directly.

A second priority is more realistic longitudinal validation. Beyond single-session or short-window multimodal settings, future evaluation should examine whether the framework can maintain coherent affective judgment across longer interaction histories, repeated engagements, and user-specific emotional trajectories. This would provide a stronger basis for assessing capabilities such as persistent personalization, long-term affective continuity, and memory-guided reinterpretation across sessions.

A third direction is capability expansion at the system level. Future versions of the Memory Bear AI infrastructure may incorporate richer modality types, more refined memory lifecycle management, and tighter integration between affective memory and downstream agent behavior. This includes not only text, speech, and vision, but also potentially interaction metadata, behavioral logs, and other signals that help characterize long-term emotional patterns in practical applications.

A fourth direction is deeper integration with real product and agent systems. The practical value of memory-centered affective judgment becomes most meaningful when it is embedded in interactive systems such as customer service platforms, educational assistants, and longer-horizon conversational agents. Future work should therefore explore how online memory updating, deployment feedback loops, and application-specific evaluation can be incorporated into the Memory Bear AI infrastructure.

Overall, the next stage of development should not be limited to improving local model performance. It should focus on building the broader memory infrastructure required for persistent affective intelligence, including datasets that can truly showcase emotional memory, more realistic longitudinal evaluation settings, and tighter connection to deployed multimodal agent systems.

\section{Conclusion}

This technical report presents the Memory Bear AI Memory Science Engine as a memory-centered architecture for multimodal affective judgment. Instead of treating affective inference as a purely local prediction problem, the proposed framework organizes emotional information through structured memory formation, short-term aggregation, long-term consolidation, retrieval, dynamic fusion calibration, and continuous updating. In this way, memory becomes a core layer for interpreting affective meaning across interaction horizons rather than an auxiliary component attached to local multimodal perception.

Overall, the report argues that the practical value of multimodal affective systems depends not only on stronger fusion, but also on the ability to preserve and reuse historically relevant emotional information. The experimental and case-based analyses suggest that this design improves long-horizon affective interpretation and robustness under noisy, incomplete, and business-grounded conditions. Although the present framework is a cognitively inspired engineering system rather than a simulation of human memory, it provides a practical path from local emotion recognition toward more continuous, stable, and deployment-relevant affective intelligence.

\newpage
\bibliography{MER_cite}
\bibliographystyle{plain}

\end{document}